\documentclass[fleqn,10pt]{wlscirep}
\usepackage[utf8]{inputenc}
\usepackage[T1]{fontenc}
\usepackage{lineno} 
\usepackage{lmodern}
\usepackage{multirow}
\usepackage{caption}
\usepackage{afterpage}
\usepackage{todonotes}
\usepackage{placeins}
\usepackage{cleveref}
\usepackage{xcolor}

\title{DALPHIN: benchmarking digital pathology AI copilots against pathologists on an open multicentric dataset}
\author[1,*]{Carlijn Lems}
\author[1]{Sander Moonemans}
\author[2,3]{Natálie Klubíčková}
\author[4]{Biagio Brattoli}
\author[4]{Taebum Lee}
\author[5]{Seokhwi Kim}
\author[6]{Veronica Vilaplana}
\author[7]{Laura Pons}
\author[7]{Sapir Hochman}
\author[7]{Mauricio Eduardo Suárez-Franck}
\author[7,8]{Pedro Luis Fernandez}
\author[9]{Julius Drachneris}
\author[9]{Donatas Petroska}
\author[9]{Renaldas Augulis}
\author[9]{Arvydas Laurinavicius}
\author[10]{Domingos Oliveira}
\author[10]{Diana Montezuma}
\author[1]{Anouk B. Bouwmeester}
\author[1]{Dominique van Midden}
\author[1]{Anne-Marie Vos}
\author[1]{Shoko Vos}
\author[1]{Jolique van Ipenburg}
\author[1,11]{Maschenka Balkenhol}
\author[1]{Koen Winkler}
\author[1]{Iris Nagtegaal}
\author[1]{Konnie Hebeda}
\author[1]{Uta Flucke}
\author[1]{Katrien Grünberg}
\author[2]{Josef Skopal}
\author[12]{Brinder S. Chohan}
\author[13]{Jordi Temprana-Salvador}
\author[14]{Enrico Munari}
\author[14]{Luca Cima}
\author[15]{Giulia Querzoli}
\author[16]{Yosamin Gonzalez Belisario}
\author[16]{Jaeike W. Faber}
\author[16]{Geert J.L.H. van Leenders}
\author[17]{Jan H. von der Thüsen}
\author[18]{Lodewijk A.A. Brosens}
\author[18,19]{Ronald R. de Krijger}
\author[19,20]{Pieter Wesseling}
\author[21,22]{Sandrine Florquin}
\author[23,24]{Mateusz Maniewski}
\author[1,23,25]{Adam Kowalewski}
\author[26,27]{Robert Barna}
\author[28,29]{Dina Tiniakos}
\author[30]{Joan Lop Gros}
\author[31]{Rogier Donders}
\author[31]{Jake S.F. Maurits}
\author[32]{Ming Yang Lu}
\author[32]{Chengkuan Chen}
\author[32]{Faisal Mahmood}
\author[1,33]{Jeroen van der Laak}
\author[1,{$\dagger$}]{Nadieh Khalili}
\author[1,{$\dagger$}]{Frédérique Meeuwsen}
\author[1]{Francesco Ciompi}

\affil[1]{Department of Pathology, Radboud University Medical Center, Nijmegen, The Netherlands}
\affil[2]{Biopticka Laboratory Ltd., Pilsen, Czech Republic}
\affil[3]{Department of Pathology, Faculty of Medicine in Pilsen, Charles University, Pilsen, Czech Republic}
\affil[4]{Lunit Inc., Seoul, South Korea}
\affil[5]{Ajou University School of Medicine, Suwon, South Korea}
\affil[6]{Universitat Politècnica de Catalunya - BarcelonaTech, Barcelona, Spain}
\affil[7]{Department of Pathology, Hospital Universitari Germans Trias i Pujol, Badalona, Spain}
\affil[8]{Faculty of Medicine and Health Sciences, Universitat Autonoma de Barcelona, Barcelona, Spain}
\affil[9]{Vilnius University and National Centre of Pathology, Vilnius, Lithuania}
\affil[10]{Research \& Development Unit, IMP Diagnostics, Porto, Portugal}
\affil[11]{Canisius Wilhelmina Ziekenhuis, Nijmegen, The Netherlands}
\affil[12]{Department of Cellular Pathology, Royal Derby Hospital, Derby, United Kingdom}
\affil[13]{Pathology Department, University Hospital Vall d'Hebron, Barcelona, Spain}
\affil[14]{Department of Diagnostic and Public Health, Pathology Unit, University and Hospital Trust of Verona, Verona, Italy}
\affil[15]{Pathology Unit, IRCCS Azienda Ospedaliero-Universitaria di Bologna, Bologna, Italy}
\affil[16]{Department of Pathology, Erasmus MC Cancer Institute, Erasmus University Medical Center, Rotterdam, The Netherlands}
\affil[17]{Department of Pathology and Clinical Bioinformatics, Erasmus University Medical Center, Rotterdam, The Netherlands}
\affil[18]{Department of Pathology, University Medical Center Utrecht, Utrecht, The Netherlands}
\affil[19]{Princess Máxima Center for Pediatric Oncology, Utrecht, The Netherlands}
\affil[20]{Department of Pathology, Amsterdam University Medical Center, VU University, Amsterdam, The Netherlands}
\affil[21]{Amsterdam UMC Location University of Amsterdam, Amsterdam, The Netherlands}
\affil[22]{Amsterdam Institute for Infection and Immunity, Amsterdam University Medical Center, Amsterdam, The Netherlands}
\affil[23]{Department of Tumor Pathology, Oncology Centre Prof. Franciszek Łukaszczyk Memorial Hospital, Bydgoszcz, Poland}
\affil[24]{Doctoral School of Medical and Health Sciences, Nicolaus Copernicus University in Toruń, Bydgoszcz, Poland}
\affil[25]{Faculty of Medicine, Bydgoszcz University of Science and Technology, Bydgoszcz, Poland}
\affil[26]{Center for Research and Innovation in Personalized Medicine of Respiratory Diseases, Victor Babes University of Medicine and Pharmacy, Timișoara, Romania}
\affil[27]{Department of Microscopic Morphology-Morphopathology, ANAPATMOL Research Center, Victor Babes University of Medicine and Pharmacy, Timișoara, Romania}
\affil[28]{Department of Pathology, Aretaieion Hospital, Medical School, National and Kapodistrian University of Athens, Athens, Greece}
\affil[29]{Translational \& Clinical Research Institute, Faculty of Medical Sciences, Newcastle University, Newcastle upon Tyne, UK}
\affil[30]{Department of Pathology, Hospital Clínic de Barcelona, Barcelona, Spain}
\affil[31]{Department IQ Health, Radboud University Medical Center, Nijmegen, The Netherlands}
\affil[32]{Department of Pathology, Brigham and Women’s Hospital, Harvard Medical School, Boston, MA, United States}
\affil[33]{Center for Medical Image Science and Visualization, Linköping University, Linköping, Sweden}
\affil[{$\dagger$}]{Shared authorship}
\affil[*]{corresponding author: Carlijn Lems (carlijn.lems@radboudumc.nl)}

\begin{abstract}
Foundation models with visual question answering capabilities for digital pathology are emerging.
Such unprecedented technology requires independent benchmarking to assess its potential in assisting pathologists in routine diagnostics.
We created DALPHIN, the first multicentric open benchmark for pathology AI copilots, comprising 1236 images from 300 cases, spanning 130 rare to common diagnoses, 6 countries, and 14 subspecialties. 
The DALPHIN design and dataset are introduced alongside a human performance benchmark of 31 pathologists from 10 countries with varying expertise.
We report results for two general-purpose (GPT-5, Gemini 2.5 Pro) and one pathology-specific copilot (PathChat+) for sequential and independent answer generation. 
We observed no statistically significant difference from expert-level performance in four of six tasks for PathChat, 2/6 tasks for Gemini, and 1/6 tasks for GPT.
DALPHIN is publicly released with sequestered, indirectly accessible ground truth to foster robust and enduring benchmarking. 
Data, methods, and the evaluation platform are accessible through \url{dalphin.grand-challenge.org}.
\end{abstract}

\begin{document}

\flushbottom
\maketitle

\thispagestyle{empty}

\section*{Introduction}

Histopathology plays a central role in the diagnosis and treatment of most cancers, inflammatory disorders, and infectious diseases.
Pathologists combine global tissue architecture with high-magnification assessment of cellular morphology in histopathology slides, relying heavily on experience-driven pattern recognition and tacit knowledge developed through years of training.
Slides are traditionally examined under a microscope and increasingly via multiresolution whole-slide images (WSIs), enabling the use of artificial intelligence (AI) tools to support routine diagnostic tasks. 

Recent advances in AI have led to the development of vision-language models (VLMs) by jointly training computers to process both natural language and digital images. 
Through \emph{instruction tuning}, i.e., using paired images and textual instructions, VLMs can be adapted for visual question answering (VQA) \cite{Anto15}. 
As a result, these models can provide relevant answers to questions posed about histopathology images, expressed as textual input prompts, either in single exchanges or across multiple turns between the user and the model.

VQA capabilities are now available in both commercial \emph{general-purpose} AI copilots (ChatGPT, Gemini) and \emph{pathology-specific} VLMs such as PathChat+ \cite{Chen25}, PRISM2 \cite{Voro25}, and SmartPath \cite{Xu25}.
These advances highlight the growing potential of AI to aid pathologists, while also raising questions about the utility and limitations of this technology in clinical diagnostics.
Addressing these questions requires quantitative \emph{benchmarks} that are independent of VLM training data and capture relevant diversity across pathology subspecialties, institutions, countries, stainings, and scanners. 

In medical AI, only a few benchmarks have been developed for VQA systems, with early efforts focusing primarily on radiology.
Since 2019, the ImageCLEF initiative \cite{Abac19} has included a VQA task for radiology and endoscopic images in its evaluation forum.
Bereuter \textit{et al.} \cite{Bere25} compared public copilots (GPT-4, Claude 3 Sonnet, Gemini 1.5) with medical students on imaging-based surgical examination questions (X-ray, CT, MRI), stratified by difficulty and modality, and found GPT-4 to outperform students, consistent with prior work \cite{Clus23}.
Similarly, the PMC-VQA dataset \cite{Zhan24} introduced a benchmark based on PubMed Central Open Access data for text-only large language models (LLMs), consisting mainly of radiology and some pathology-related questions.

Some initiatives have aimed to establish pathology-specific VQA benchmarks. 
PathVQA \cite{He20} was designed to test whether AI could pass the board-certified examination of the American Board of Pathology, but has two main limitations: 1) lack of pathologist review, resulting in inclusion of non-pathology images; and 2) full public availability, enabling potential leakage of test data into VQA training sets.
The recent PathQABench introduced in the PathChat study \cite{Lu24} partly addresses these issues with a pathologist-curated dataset split into public and private subsets.
However, its \emph{public} subset is based on the largely adopted TCGA archive, while its \emph{private} subset is not accessible, preventing result reproduction or benchmarking of other models.
This is particularly relevant given the rapid evolution of models, with newer versions often released by the time studies are published.
Recent non-VQA pathology benchmarks \cite{Steg26,Rijt26,Bult22} address these issues via sequestered evaluation on a trusted third-party platform, where models are run on-platform without releasing test data.
However, this precludes benchmarking commercial models that cannot be uploaded to such a platform.
Finally, most studies lack comparison with pathologists across different expertise levels, limiting the clinical interpretability of model performance.

\begin{figure}[htbp]
\centering
\includegraphics[width=1.0\textwidth]{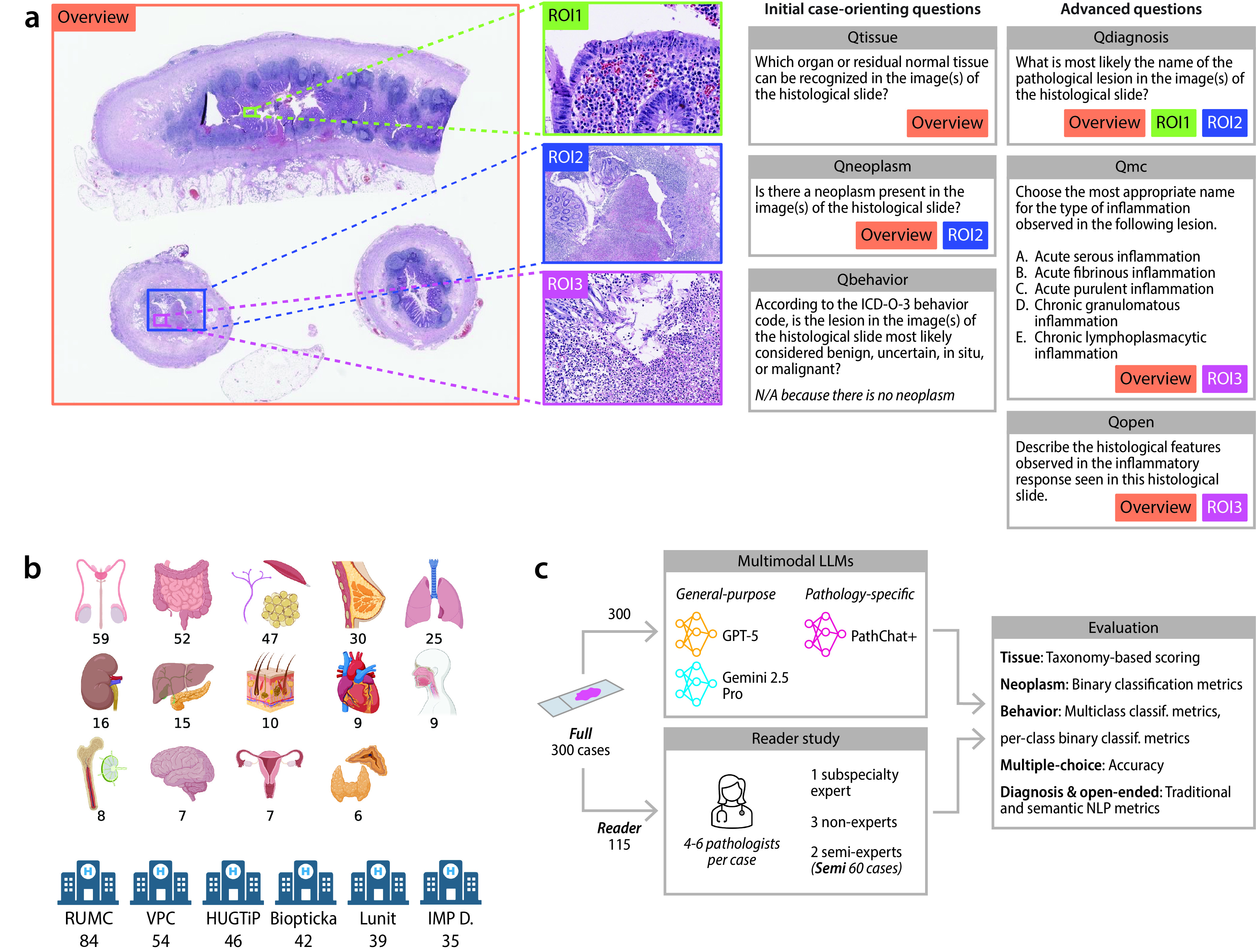}
\caption{Overview of this study.
(a) Illustrative example demonstrating the format of a DALPHIN benchmark case. Each case includes a low-resolution overview of one or more histopathology whole-slide images and one or more regions of interest (ROIs) selected by the contributing pathologist. In addition, each case includes four standard questions (when applicable), a case-specific multiple-choice question, and optionally a case-specific free-response question. The contributing pathologist provides the reference-standard answers to all questions, formulates the case-specific questions, and determines which ROIs are supplied as input for each question.
(b) Composition of DALPHIN. The benchmark comprises 300 cases spanning 14 subspecialties - genitourinary, gastrointestinal, soft tissue, breast, thoracic, nephropathology, hepatopancreatobiliary, dermatopathology, cardiovascular, head and neck, hematopathology, neuropathology, gynecologic pathology, and endocrine (ordered from top left to bottom right in the figure) - collected from six healthcare centers across six countries.
RUMC: Radboud University Medical Center; VPC: Vilnius University and National Centre of Pathology; HUGTiP: Hospital Universitari Germans Trias i Pujol; IMP D.: IMP Diagnostics.
(c) Workflow for evaluating VLMs on DALPHIN, collecting responses from pathologists of different expertise levels in a reader study, and comparing all responses to the reference-standard answers.
Created in Biorender. Lems, C. (2026) \url{https://BioRender.com/oozx89q}.
}
\label{fig:workflow}
\end{figure}

In this study, we introduce the digital pathology AI copilot benchmark (DALPHIN), a multicentric open benchmark for evaluating digital pathology AI copilots (\cref{fig:workflow}).
The dataset comprises 300 cases from six international healthcare institutions, covering 130 diagnoses across 14 pathology subspecialties, including neoplastic, non-neoplastic, and rare entities.
DALPHIN contains 1236 histology images, provided as low-resolution WSIs and higher-resolution regions of interest (ROIs) extracted from hematoxylin and eosin (H\&E) or periodic acid-Schiff (PAS) stained WSIs.
This mimics histopathological assessment, where pathologists alternate between low-magnification evaluation of overall tissue context and high-magnification inspection of diagnostically relevant detail.
Each case includes up to six questions, with answers provided by the case-contributing pathologist serving as the reference standard (see `Benchmark' in Methods).
Apart from organ or tissue type in most questions, no standardized clinical context is provided.

Because diagnostic interpretation in pathology is subject to interobserver variability, even expert performance should be interpreted in this context, as disagreement with the reference standard may reduce measured scores.
To place model performance in a clinically relevant context, we compare VLMs with pathologists across different expertise levels.
We conducted a reader study with 31 pathologists, including 24 board-certified pathologists and seven pathology residents, stratified into subspecialty experts, semi-experts, and non-experts (see `Reader study' in Methods).

We publicly release DALPHIN images and questions while keeping reference answers sequestered and indirectly accessible via a secure evaluation procedure that processes answers submitted to the Grand Challenge platform \cite{Meak25}.
This prevents leakage into AI model training sets while enabling benchmarking of both local and online AI copilots, now and in the future.

\section*{Results}

To assess their capabilities as digital pathology copilots, we evaluated two general-purpose VLMs (Gemini 2.5 Pro \cite{Goog25}, GPT-5 \cite{Open25}) and one pathology-focused VLM (PathChat+) on DALPHIN.
To simulate a more realistic diagnostic workflow, models and pathologists answered each case's questions \emph{sequentially}, with prior context informing subsequent responses (see `Reader study' and `VLM evaluation' in Methods).
Full pathologist review of all 300 cases across expertise levels was not feasible due to the required number of readers and per-reader case burden.
We therefore defined three nested dataset subsets with increasing levels of pathologist evaluation: 
(1) \emph{DALPHIN$_{full}$} ($n=300$), the complete dataset; (2) \emph{DALPHIN$_{reader}$} ($n=115$), a subset of DALPHIN$_{full}$ in which each case was reviewed by one subspecialty expert and three non-expert (resident) pathologists; and 
(3) \emph{DALPHIN$_{semi}$} ($n=60$), a subset of DALPHIN$_{reader}$ spanning seven of 14 subspecialties, in which each case was additionally reviewed by two semi-expert pathologists.
An example case with responses from all VLMs and expertise levels is shown in Extended Data \cref{fig:example-case}.
Exact $P$-values are provided in Supplementary \cref{tab:p_values}.
Subgroup analyses (e.g., by subspecialty) were exploratory, and no $P$-values are reported.

\subsection*{Performance on initial case-orienting questions}

\begin{figure}[htbp]
\centering
\includegraphics[width=1.0\textwidth]{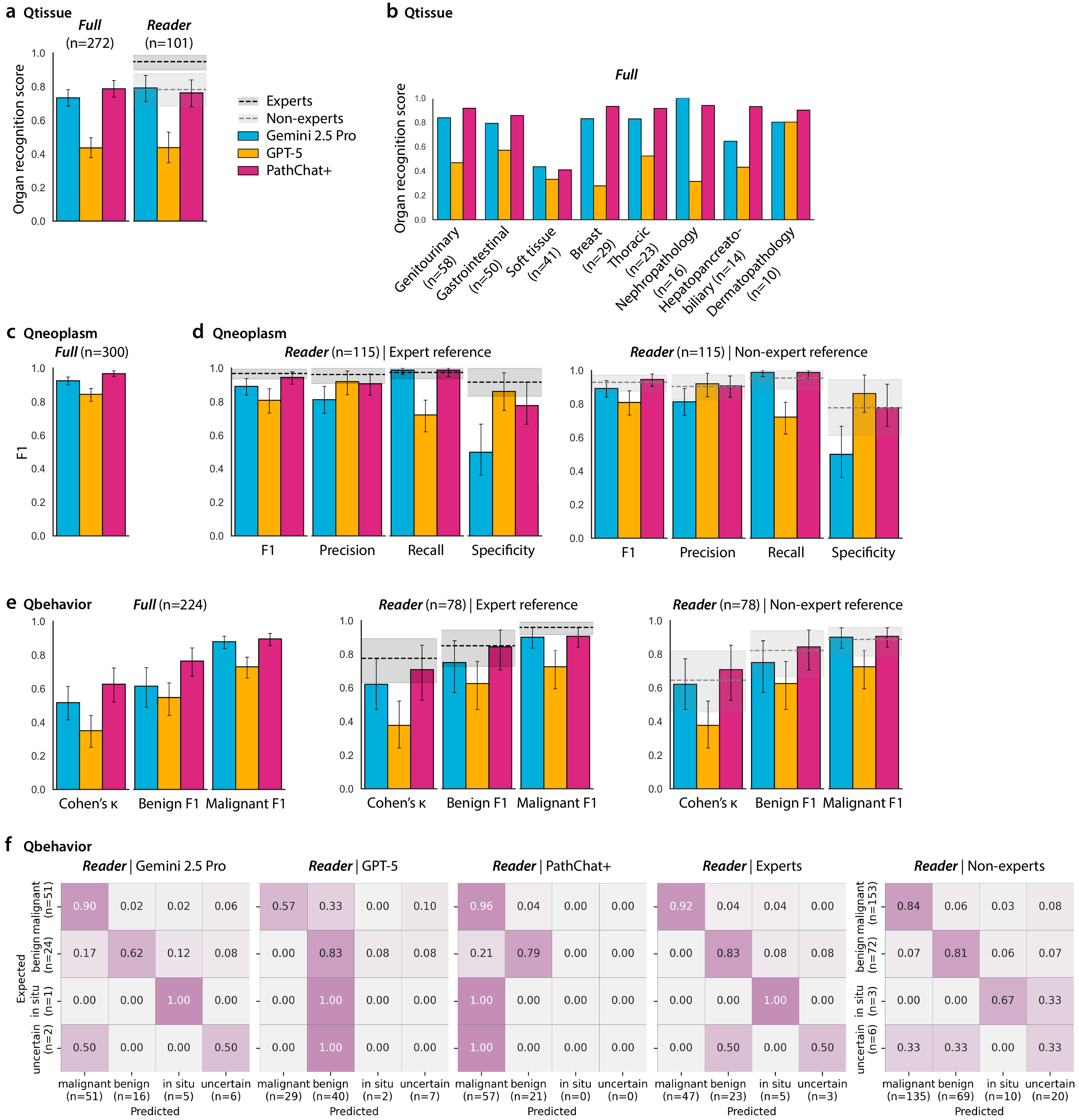}
\caption{Evaluation of VLMs on initial case-orienting tasks and comparison with subspecialty expert and non-expert (resident) pathologists.
Error bars for VLMs and shaded regions for pathologists indicate 95\% confidence intervals.
(a) Organ-recognition performance of VLMs, experts, and non-experts on a free-response organ recognition task ($Q_{\text{tissue}}$) in DALPHIN$_{full}$ and DALPHIN$_{reader}$.
(b) Organ-recognition performance of VLMs in DALPHIN$_{full}$ stratified by pathology subspecialty; only subspecialties with $\geq$10 cases are shown.
(c) F1 scores of VLMs for neoplastic status ($Q_{\text{neoplasm}}$) in DALPHIN$_{full}$.
(d) Binary classification metrics for $Q_{\text{neoplasm}}$ for VLMs, experts, and non-experts in DALPHIN$_{reader}$.
(e) Multiclass classification metrics for neoplastic behavior (benign, uncertain, in situ, malignant) ($Q_{\text{behavior}}$) for VLMs, experts, and non-experts in DALPHIN$_{full}$ and DALPHIN$_{reader}$.
(f) Confusion matrices for $Q_{\text{behavior}}$ for VLMs, experts, and non-experts in DALPHIN$_{reader}$ ($n=78$). Counts for non-experts are multiplied by three because each case was assessed by three independent non-experts.
}
\label{orienting-results}
\end{figure}

\subsubsection*{Gemini and PathChat surpass GPT in organ recognition, but experts remain superior}

We first assessed performance on `initial orienting questions' during case review, starting with organ or residual normal tissue identification ($Q_{\text{tissue}}$).
Although clinicians typically provide the specimen origin, pathologists commonly verify that it matches the image as a quality check to exclude specimen mix-ups.
For 272 of 300 cases ($90.7\%$), the tissue type was identifiable from the images, and responses from VLMs and pathologists were evaluated against the ground truth using a taxonomy-based hierarchical scoring system (1.0 for exact matches; 0.75/0.5 for one/two steps away; else 0; see `Evaluation' in Methods), referred to as the organ recognition score.

Focusing on DALPHIN$_{full}$, Gemini and PathChat outperformed GPT in terms of organ recognition score ($+68.6\%$ and $+80.6\%$, respectively), with PathChat also outperforming Gemini ($+7.1\%$) (\cref{orienting-results}a and Supplementary \cref{tab:q_tissue}).
In DALPHIN$_{reader}$, experts outperformed all three VLMs ($+19.7\%$ vs Gemini, $+116.4\%$ vs GPT, $+24.4\%$ vs PathChat).
Gemini and PathChat showed no statistically significant differences compared with non-experts in DALPHIN$_{reader}$ or with semi-experts in DALPHIN$_{semi}$, whereas GPT scored below all expertise levels ($-44.0\%$ relative to non-experts in DALPHIN$_{reader}$; $-41.3\%$ relative to semi-experts in DALPHIN$_{semi}$).
Together, these findings suggest that Gemini and PathChat outperform GPT in organ recognition but remain below expert pathologists.

Analysis of DALPHIN$_{full}$ further revealed variation in VLM performance across subspecialties (\cref{orienting-results}b and Supplementary \cref{tab:q_tissue_subspecialty}).
Skin was consistently well-recognized by all VLMs (mean score: $0.80$--$0.90$), whereas soft tissue was most challenging ($0.33$--$0.43$).
The largest performance margins of PathChat and Gemini over GPT were observed in breast and nephropathology.
Notably, PathChat did not provide a tissue prediction for 36 cases, instead indicating the absence of residual normal tissue.
These omissions spanned 11 of 14 subspecialties and were most frequent in soft tissue pathology (17/41 cases, $41.5\%$).
Overall, subspecialty analysis reveals heterogeneous performance across domains.

\subsubsection*{Gemini overcalls neoplasms while GPT undercalls them}

Next, we evaluated neoplastic versus non-neoplastic classification ($Q_{\text{neoplasm}}$).
For this and subsequent questions, the tissue type was provided, reflecting standard clinical practice.
Performance was assessed using binary classification metrics (F1, Matthews Correlation Coefficient [MCC], precision, recall, specificity).

In DALPHIN$_{full}$, PathChat outperformed Gemini (F1 $+4.6\%$, MCC $+37.8\%$) and GPT (F1 $+14.7\%$, MCC $+58.5\%$) in overall classification, while Gemini significantly exceeded GPT only in F1 ($+9.6\%$) (\cref{orienting-results}c-d and Supplementary \cref{tab:q_neoplasm}).
In DALPHIN$_{reader}$, PathChat scored between expert and non-expert pathologists and showed no significant differences compared with either group, whereas Gemini was outperformed by experts (F1 $-8.0\%$, MCC $-32.3\%$), and GPT scored below non-experts (F1 $-12.9\%$, MCC $-28.6\%$).
Consistently, GPT was also surpassed by semi-experts in DALPHIN$_{semi}$ (F1 $-18.8\%$, MCC $-51.0\%$).

In further analysis of DALPHIN$_{reader}$, PathChat showed expert recall, with precision and specificity comparable to non-experts and not significantly different from experts (\cref{orienting-results}d).
Gemini also demonstrated expert-level recall but scored below non-experts in precision ($-10.1\%$) and specificity ($-35.7\%$), indicating a tendency to overcall neoplasms.
GPT showed the opposite pattern, approaching expert-level specificity and precision while being surpassed by non-experts in recall ($-24.3\%$).
Taken together, Gemini tends to overcall neoplasia, whereas GPT is more conservative and more likely to miss neoplastic entities.

Lastly, seven non-expert responses were not `yes' or `no' (all `unsure' or similar) and were handled via probabilistic imputation (see `Reader study' in Methods).
To test robustness to these assumptions, we additionally evaluated best- and worst-case scenarios in which responses were forced to match or not match the reference answer (Supplementary \cref{tab:q_neoplasm_imputation}).
Across metrics, estimates differed by $\sim0$--$7\%$ (e.g., F1: $0.922$--$0.936$).

\subsubsection*{Gemini and PathChat more often misclassify benign lesions as malignant, whereas GPT tends to misclassify malignant lesions as benign}

The last orienting question assessed neoplastic behavior classification (benign, uncertain, in situ, malignant) following the International Classification of Diseases for Oncology, 3rd Edition (ICD-O-3) ($Q_{\text{behavior}}$).
Because this applied only to neoplastic cases (224/300, $74.7\%$) and sample size in DALPHIN$_{semi}$ was limited, analyses were restricted to DALPHIN$_{full}$ and DALPHIN$_{reader}$.
Class-specific metrics focused on benign and malignant lesions, which are more prevalent and better represented in our dataset, whereas uncertain ($n=7$) and in situ ($n=3$) cases were too rare for meaningful evaluation.
Performance was assessed using overall metrics (Cohen's $\kappa$, MCC) and class-specific F1 and MCC.

In DALPHIN$_{full}$, PathChat outperformed Gemini in benign classification (F1 $+24.2\%$, MCC $+35.5\%$) and GPT across all metrics (range of $+22.5\%$ to $+85.2\%$) (\cref{orienting-results}e and Supplementary \cref{tab:q_behavior}).
Gemini, in turn, outperformed GPT in overall (Cohen's $\kappa$ $+47.5\%$, MCC $+33.9\%$) and malignant classification (F1 $+20.2\%$, MCC $+37.9\%$).
In DALPHIN$_{reader}$, no significant differences were observed between PathChat, Gemini, and experts or non-experts, although PathChat generally ranged from expert- to non-expert-level performance, while Gemini mostly scored at or below the non-expert level.
GPT performed significantly worse than experts across all metrics (range of $-51.4\%$ to $-24.4\%$) and worse than non-experts on all but malignant MCC ($-42.8\%$ to $-18.3\%$).
Overall, PathChat performs best, followed by Gemini and then GPT.

Confusion matrices for DALPHIN$_{full}$ (Extended Data \cref{fig:cm-full}) show that PathChat and Gemini more often misclassified benign lesions as malignant than vice versa (PathChat: $16.1\%$ vs $10.8\%$; Gemini: $19.6\%$ vs $8.9\%$), whereas GPT showed the opposite pattern ($10.7\%$ vs $33.5\%$).
In DALPHIN$_{reader}$ (\cref{orienting-results}f), all VLMs more frequently confused benign and malignant lesions with one another than expert and non-expert pathologists, who more often misclassified them as in situ or uncertain rather than benign or malignant.
Finally, VLMs differed in their use of the in situ and uncertain labels: in DALPHIN$_{full}$, Gemini produced the most such predictions (13 in situ, 16 uncertain), followed by GPT (6, 15), while PathChat never predicted these classes.

Similar to $Q_{\text{neoplasm}}$, two expert (`premalignant') and two non-expert responses (`unsure', `I don't know') fell outside the predefined categories and were handled via probabilistic imputation (see `Reader study' in Methods).
Best- and worst-case scenarios were also evaluated by forcing responses to match or not match the reference answer (Supplementary \cref{tab:q_behavior_imputation}).
Across metrics, estimates varied by $\sim0$--$8\%$ (e.g., Cohen’s $\kappa$ for experts: $0.772$--$0.822$).

\subsection*{Performance on advanced questions}

We next evaluated performance on `advanced questions', including a standardized free-response diagnostic question ($Q_{\text{diagnosis}}$), a case-specific multiple-choice question ($Q_{\text{mc}}$), and, for most cases (285/300, $95.0\%$), a case-specific free-response question ($Q_{\text{open}}$).
Both case-specific questions were defined by the contributing pathologist.

\begin{figure}
\centering
\includegraphics[width=1.0\textwidth]{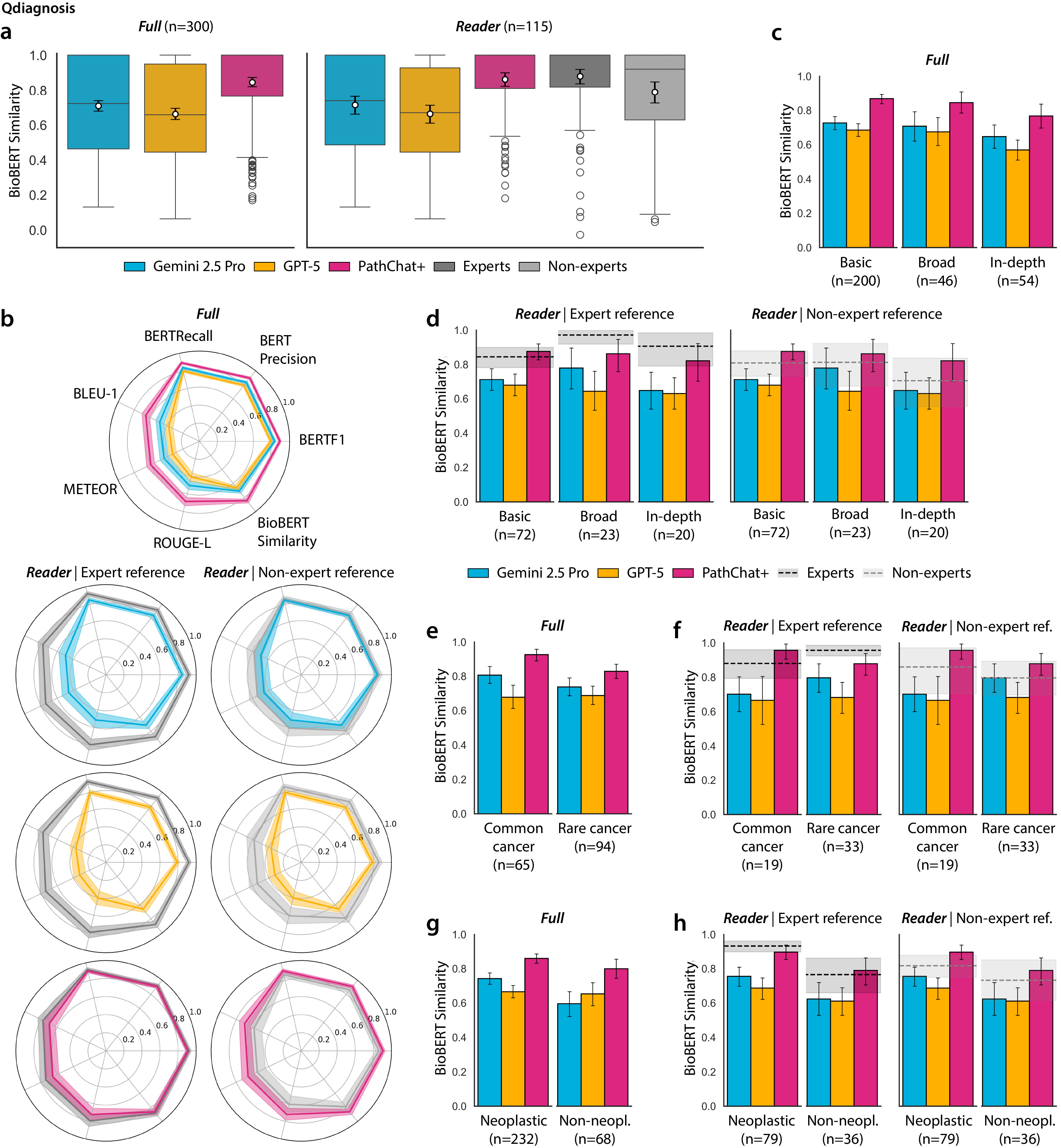}
\caption{Evaluation of VLMs on a free-response diagnosis task ($Q_{\text{diagnosis}}$) and comparison with subspecialty expert and non-expert (resident) pathologists.
(a) BioBERT similarity scores of VLMs, experts, and non-experts in DALPHIN$_{full}$ and DALPHIN$_{reader}$. White points with error bars indicate mean scores with 95\% confidence intervals.
(b) Traditional and semantic NLP metric scores for VLMs, experts, and non-experts, shown with mean values and 95\% confidence intervals. CIDEr is excluded to preserve visibility of the other metrics.
(c-h) BioBERT similarity scores for VLMs, experts, and non-experts across diagnostic subsets: basic/broad/in-depth knowledge per Dutch pathology education guidelines (c–d), common/rare cancers (e-f), and neoplastic/non-neoplastic entities (g-h). Panels c, e, and g show DALPHIN$_{full}$; panels d, f, and h show DALPHIN$_{reader}$. Error bars for VLMs and shaded regions for pathologists indicate 95\% confidence intervals.
}
\label{fig:diagnosis-results}
\end{figure}

\subsubsection*{PathChat outperforms general-purpose models in diagnosing lesions, approaching expert-level performance}

$Q_{\text{diagnosis}}$ tested the ability to provide the most likely diagnosis without any clinical context beyond the tissue type.
Model and pathologist responses were evaluated against the ground truth and synonyms using natural language processing (NLP) overlap metrics (BLEU-1, ROUGE-L, CIDEr, METEOR; measuring textual overlap) and semantic similarity metrics (BERTScore precision, recall, F1; BioBERT similarity; measuring similarity in meaning).

In DALPHIN$_{full}$, PathChat outperformed Gemini (range of $+6.5\%$ to $+41.9\%$) and GPT ($+11.6\%$ to $+81.9\%$) across all metrics, with Gemini exceeding GPT ($+4.8\%$ to $+32.0\%$) (\cref{fig:diagnosis-results}a-b and Supplementary \cref{tab:q_diagnosis_sem,tab:q_diagnosis_trad}).
When compared with experts or non-experts in DALPHIN$_{reader}$ and semi-experts in DALPHIN$_{semi}$, GPT scored below all expertise levels ($-56.3\%$ to $-7.4\%$).
Gemini scored significantly below experts across all metrics ($-38.1\%$ to $-7.7\%$), below semi-experts for four metrics ($-27.4\%$ to $-12.4\%$), and below non-experts only on BioBERT similarity ($-9.7\%$).
PathChat outperformed non-experts on all metrics ($+5.1\%$ to $+23.0\%$) and showed no significant differences from expert or semi-expert scores.
Overall, PathChat outperforms general-purpose models in diagnosis, approaching expert-level performance.

We stratified cases along four axes: diagnostic difficulty (basic [A], broad [B], in-depth knowledge [C]; Dutch pathology education guidelines \cite{NVP24}), cancer incidence (rare cancers <6 per 100,000 per year \cite{Gatt11}), neoplastic status, and subspecialty.
Across the first three axes, PathChat consistently achieved higher average scores than the general-purpose models in DALPHIN$_{full}$, suggesting its superiority is not domain-specific (\cref{fig:diagnosis-results}c-h and Supplementary \cref{tab:q_diagnosis_difficulty_full_sem,tab:q_diagnosis_difficulty_full_trad,tab:q_diagnosis_difficulty_reader_sem,tab:q_diagnosis_difficulty_reader_trad,tab:q_diagnosis_cancer_incidence_full_sem,tab:q_diagnosis_cancer_incidence_full_trad,tab:q_diagnosis_cancer_incidence_reader_sem,tab:q_diagnosis_cancer_incidence_reader_trad,tab:q_diagnosis_neoplastic_status_full_sem,tab:q_diagnosis_neoplastic_status_full_trad,tab:q_diagnosis_neoplastic_status_reader_sem,tab:q_diagnosis_neoplastic_status_reader_trad}).
Performance varied by subspecialty, with PathChat generally scoring highest and Gemini and GPT showing inconsistent relative rankings (Extended Data \cref{fig:diagnosis-subspecialty} and Supplementary \cref{tab:q_diagnosis_subspecialty_sem,tab:q_diagnosis_subspecialty_trad}).
In DALPHIN$_{reader}$, expert pathologists showed a slight advantage over PathChat for broad and in-depth diagnoses and rare cancers, while the general-purpose models showed lower performance than non-experts for basic diagnoses and common cancers.
Taken together, PathChat outperforms general-purpose models and approaches expert-level performance across several diagnostic strata.

\subsubsection*{General-purpose and pathology-specific models perform similarly on case-specific multiple-choice VQA}

$Q_{\text{mc}}$ questions assessed five domains of pathology expertise: morphology, ancillary testing, diagnosis, clinical knowledge, and mixed reasoning (see `Benchmark' in Methods).
Gemini and GPT did not answer one actinic keratosis question, deeming it inapplicable after misdiagnosing the lesion as a different entity.

\begin{figure}
\centering
\includegraphics[width=0.9\textwidth]{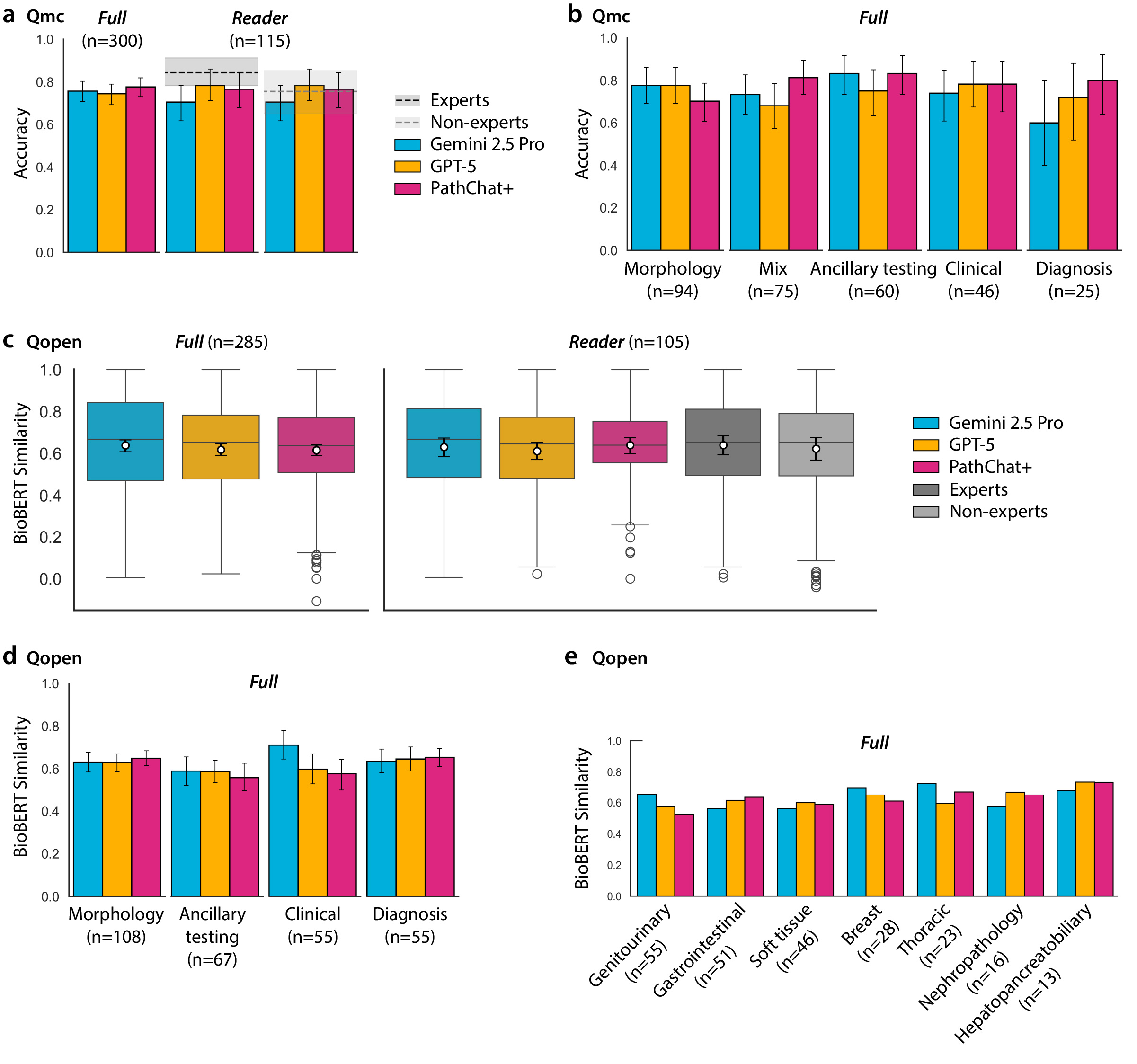}
\caption{Evaluation of VLMs on case-specific multiple-choice ($Q_{\text{mc}}$) and free-response questions ($Q_{\text{open}}$) and comparison with subspecialty expert and non-expert (resident) pathologists.
Unless otherwise noted, error bars for VLMs and shaded regions for pathologists indicate 95\% confidence intervals.
(a) Accuracy of VLMs, experts, and non-experts on $Q_{\text{mc}}$ questions in DALPHIN$_{full}$ and DALPHIN$_{reader}$.
(b) Accuracy of VLMs for $Q_{\text{mc}}$ in DALPHIN$_{full}$ stratified by question category.
(c) BioBERT similarity scores on $Q_{\text{open}}$ questions for VLMs, experts, and non-experts in DALPHIN$_{full}$ and DALPHIN$_{reader}$. White points with error bars indicate mean scores with 95\% confidence intervals.
(d-e) BioBERT similarity scores of VLMs for $Q_{\text{open}}$ in DALPHIN$_{full}$ stratified by question category (d) and by pathology subspecialty (e); only subspecialties with $\geq$10 cases are shown.
}
\label{fig:case-specific-results}
\end{figure}

No significant differences in accuracy were observed among VLMs in DALPHIN$_{full}$ (accuracy range of $74.3\%$ to $77.7\%$) (\cref{fig:case-specific-results}a and Supplementary \cref{tab:q_mc}).
In DALPHIN$_{reader}$, experts surpassed Gemini ($+19.8\%$), with no other significant differences between VLMs and experts or non-experts in DALPHIN$_{reader}$, or semi-experts in DALPHIN$_{semi}$.
Still, GPT and PathChat were closer to non-expert than expert performance in DALPHIN$_{reader}$.
In-depth analysis of DALPHIN$_{full}$ showed some variation across question categories (\cref{fig:case-specific-results}b and Supplementary \cref{tab:q_mc_category}).
Overall, all VLMs exhibit similar performance in multiple-choice VQA.

\subsubsection*{Gemini shows similar or higher performance than GPT and PathChat in case-specific free-response VQA}

$Q_{\text{open}}$ questions spanned the same domains as $Q_{\text{mc}}$ except mixed and were evaluated using NLP metrics.
Response length in characters varied across VLMs: Gemini was the shortest (mean 93, median 39), followed by GPT (110, 76), and PathChat was the longest (128, 81).
PathChat did not answer two questions regarding ancillary testing and macroscopic appearance, indicating that the information was unavailable.

In DALPHIN$_{full}$, Gemini outperformed PathChat on seven metrics (range of $+3.3\%$ to $+35.9\%$) and GPT on six ($+4.8\%$ to $+43.7\%$) (\cref{fig:case-specific-results}c and Supplementary \cref{tab:q_open_sem,tab:q_open_trad}).
In DALPHIN$_{reader}$ and DALPHIN$_{semi}$, Gemini showed no significant differences with any pathology expertise level.
PathChat was outperformed by experts on BERTScore precision ($-10.5\%$) and BERTScore F1 ($-5.7\%$), and by non-experts only on BERTScore precision ($-7.6\%$).
GPT was surpassed by all expertise levels on BERTScore precision (range of $-11.8\%$ to $-8.9\%$) and by experts on BERTScore F1 ($-4.9\%$) and METEOR ($-21.8\%$), but exceeded semi-experts on BERTScore recall ($+6.9\%$).
Stratification of DALPHIN$_{full}$ suggests that Gemini’s higher overall performance is driven largely by the clinical category (\cref{fig:case-specific-results}d and Supplementary \cref{tab:q_open_category_sem,tab:q_open_category_trad}) and by genitourinary, breast, and thoracic cases (\cref{fig:case-specific-results}e and Supplementary \cref{tab:q_open_subspecialty_sem,tab:q_open_subspecialty_trad}).
In sum, Gemini's performance advantage in DALPHIN$_{full}$ appears driven by specific categories and subspecialties, while model–pathologist differences are minimal in DALPHIN$_{reader}$ and DALPHIN$_{semi}$.

\subsection*{Contextual versus independent VLM evaluation}

In the results above, models answered all questions sequentially per case, retaining prior context from the same case, referred to as the \emph{contextual} (feedback) generation scenario (see `Answer generation scenarios' in Methods).
This mimics a pathologist’s follow-up workflow but allows early errors to propagate to later responses.
We therefore also evaluated an \emph{independent} generation scenario, where each question is treated as a standalone query.

When comparing VLM performance across contextual and independent generation on DALPHIN$_{full}$, most metrics showed no significant differences (Supplementary \cref{tab:q_all_i_contextual_independent,tab:q_all_ii_contextual_independent,tab:q_all_iii_contextual_independent}).
PathChat performed significantly better in the contextual scenario for most $Q_{\text{neoplasm}}$ metrics, and GPT for three of six $Q_{\text{behavior}}$ metrics.
For $Q_{\text{mc}}$, PathChat showed a similar pattern, whereas the general-purpose models achieved significantly higher accuracy in the independent scenario.
Overall, performance was largely comparable across scenarios, suggesting a trade-off between the benefits of prior case context and the risk of error propagation.

\section*{Discussion}

The rapid development of AI copilots has been accompanied by growing efforts to evaluate their performance across domains.
While benchmarking is well established in computer vision and natural language processing, systematic evaluation of AI copilots in pathology remains limited, reflecting their more recent emergence in this field.
Here, we introduce DALPHIN, the first multicentric open benchmark for digital pathology copilots. 

DALPHIN provides an open evaluation framework enabling transparent, continuous reassessment as new models emerge. 
Given the rapid evolution of VLM-based copilots, model versions change quickly. 
By sequestering the ground truth and making it indirectly accessible, DALPHIN enables robust, long-term evaluation, mitigates the risk of data leakage into training sets, and standardizes benchmarking of current and future pathology AI copilots.

Beyond providing a benchmarking framework, we evaluate several state-of-the-art VLMs against pathologists of varying expertise.
The pathology-specific model outperformed general-purpose models on several tasks, underscoring the value of domain-specific training.
However, performance patterns varied considerably across tasks, suggesting that no single metric fully reflects model performance and highlighting the need for multitask benchmarking, as implemented in DALPHIN.

Despite variability across tasks, VLMs demonstrated consistent model-specific behavior across classification tasks ($Q_{\text{neoplasm}}$, $Q_{\text{behavior}}$).
Gemini exhibited a more sensitivity-oriented tendency, classifying more cases as neoplastic or malignant, whereas GPT showed the opposite pattern.
The limited transparency of general-purpose model training makes it difficult to pinpoint the underlying causes, which may stem from differences in training data or model calibration.
Nevertheless, these behavioral patterns may be relevant when considering potential model use.

A potential application of AI copilots is supporting rare cancer diagnosis, particularly in settings lacking subspecialty expertise, such as smaller hospitals or resource-constrained regions. In our limited dataset, none of the models reached expert-level performance on rare cancers. This gap may reflect the scarcity and heterogeneity of rare cancers in large-scale datasets, which may limit robust pattern generalization. 
 
DALPHIN offers unique insights into models' ability to reason sequentially by comparing independent and contextual evaluations.
The sequential setting mimics clinical practice, where pathologists progress from basic observations such as tissue identification to more complex diagnostic questions. 
While contextual information can support consistency and stepwise reasoning, it may also introduce anchoring effects, where early impressions influence subsequent judgments, as in routine clinical practice.
Early misclassifications may therefore propagate through later responses.
Consistent with this, GPT and Gemini showed lower performance on later-stage $Q_{\text{mc}}$ questions in the sequential setting. 
In one actinic keratosis case, both models declined to answer due to disagreement with the preceding diagnosis. 
This illustrates how prior context can affect downstream responses in sequential reasoning, though this interpretation remains exploratory.

Several limitations should be considered when interpreting the results of DALPHIN. 
Questions were designed to be answerable from selected ROIs from H\&E or PAS slides, as most current VLMs cannot directly process WSIs, a key constraint for their application in digital pathology. 
Although DALPHIN provides WSI-level information via overview images, this is limited by low resolution. 
While the restriction to ROIs enables benchmarking of existing models, routine clinical diagnosis relies on full WSI examination, integration of clinical information, and, in many cases, ancillary investigations such as immunohistochemistry (IHC) or molecular testing. 
Accordingly, the benchmark captures only a subset of diagnostic information. 
Prior work has shown that including clinical context alongside pathology images can improve diagnostic VQA performance \cite{Lu24}. 
Future extensions could incorporate full WSIs, clinical data, and IHC to better reflect real-world diagnostic workflows. 

A further limitation relates to dataset composition: DALPHIN includes relatively few entities with uncertain or in situ behavior, which are potentially more diagnostically challenging. 
Consistent with this, PathChat did not correctly classify the behavior of any such cases in the present study. 
Future work could increase their representation in the benchmark.

Another limitation concerns the evaluation of free-response diagnostic and case-specific answers ($Q_{\text{diagnosis}}$, $Q_{\text{open}}$), which relied on standard NLP metrics. 
These metrics are based on semantic similarity and may not capture histopathological nuance, where different formulations can be clinically equivalent, partially correct, or conceptually related. 
Response style may also influence scoring; for example, in $Q_{\text{open}}$, PathChat generated longer explanations, reducing BERTScore precision, whereas more concise responses may achieve higher similarity scores despite comparable reasoning. 
Future work could explore alternative evaluation strategies, including LLM-based judges to assess clinical correctness or reasoning quality, though this introduces model-dependent bias. 
Another direction is the development of pathologist-defined metrics aimed at capturing clinically meaningful correctness.
While beyond the scope of this study, improving scalable and clinically grounded evaluation remains an important direction for future research.

In summary, DALPHIN provides an open, robust framework for evaluating pathology AI copilots, enabling assessment of rapidly evolving VLMs on clinically relevant tasks against pathologists of varying expertise. 
Results demonstrate substantial task-dependent variability across models and question types. 
By enabling independent evaluation of newly released models, DALPHIN promotes a more transparent and continuously evolving understanding of model capabilities, supporting careful consideration of their potential clinical use.

\section*{Online Methods}

\subsection*{Benchmark}

\subsubsection*{Case collection, preparation, and digitization}

We collected 300 cases from six healthcare institutions: the Radboud University Medical Center (RUMC, Nijmegen, the Netherlands), the Vilnius University and National Centre of Pathology (VPC, Vilnius, Lithuania), the Hospital Universitari Germans Trias i Pujol (HUGTiP, Badalona, Spain), the Biopticka Laboratory Ltd. (Pilsen, Czech Republic), Lunit (Seoul, South Korea) but originating from Ajou University Medical Center (AUMC, Suwon, South Korea), and IMP Diagnostics (Porto, Portugal).
All slides were prepared and digitized in the originating laboratory using one of five scanners: 3DHistech Pannoramic 1000 (RUMC, HUGTiP), 3DHistech Pannoramic 250 (Biopticka), Hamamatsu NanoZoomer S360 (Biopticka), Leica Biosystems Aperio GT450 (Biopticka, IMP Diagnostics), and Leica Biosystems Aperio AT2 (VPC, Lunit).
All images were converted to a standard tagged image file format (TIFF).

\subsubsection*{Case selection}

In total, 14 board-certified pathologists and two pathology residents selected cases spanning 14 subspecialties and 130 diagnoses, including 41 non-neoplastic entities and a range of diagnostic difficulty.
Cases were included only if a final diagnosis could be established based on hematoxylin and eosin (H\&E) and/or periodic acid–Schiff (PAS) slides alone.
Although ancillary tests or clinical data may be required for formal confirmation in some entities, only morphologically characteristic cases were selected.

\subsubsection*{Questions and answers}

Each case included up to six questions with reference answers from the contributing pathologist.
These answers served as the reference standard, as the contributing pathologist had access to the full whole-slide image(s), clinical context, and ancillary test results.

\paragraph{Case-orienting questions}
For each case, we included three initial \emph{case-orienting} questions: ($Q_{\text{tissue}}$) \textit{`Which organ or residual normal tissue can be recognized in the image(s) of the histological slide?'}; ($Q_{\text{neoplasm}}$) \textit{`Is there a neoplasm present in the image(s) of the histological slide?'}; and ($Q_{\text{behavior}}$) \textit{`According to the ICD-O-3 behavior code (benign: 0, uncertain: 1, in situ: 2, malignant: 3/6/9), is the lesion in the image(s) of the histological slide most likely considered benign, uncertain, in situ, or malignant?'}.
$Q_{\text{tissue}}$ was considered not applicable when no recognizable organ or residual normal tissue was present (e.g., neuropathology cases lacking preexisting brain tissue) (28 of 300 cases).
$Q_{\text{behavior}}$, applicable only to neoplastic lesions (232 of 300 cases), followed the International Classification of Diseases for Oncology, 3rd Edition (ICD-O-3); malignant codes /3 (primary), /6 (metastatic), and /9 (uncertain primary or metastatic) were grouped for analysis.
For eight neoplastic cases, ICD-O-3 behavior could not be assigned based on H\&E morphology alone (e.g., gastrointestinal stromal tumors); therefore, $Q_{\text{behavior}}$ was deemed non-applicable to these cases, leaving 224 cases for analysis.

\paragraph{Advanced questions}
For each case, we also included two or three \emph{advanced} questions.
The first ($Q_{\text{diagnosis}}$) asked \textit{`What is most likely the name of the lesion in the image(s) of the histological slide? In case it is a neoplastic lesion, please name the lesion according to the WHO Classification of Tumors, if possible'}.
The remaining one or two questions were case-specific: a required multiple-choice question ($Q_{\text{mc}}$) with three to six answer options and, for 285 of 300 cases, an additional open-ended question ($Q_{\text{open}}$) (15 were excluded; see `Data curation').
Both questions were provided by the contributing pathologist.
A small subset of $Q_{\text{mc}}$ items allowed multiple correct answers.
Pathologists were encouraged to focus questions on the lesion shown in the image(s), avoiding general knowledge, and to phrase them to elicit concise, precise responses suitable for automated evaluation with similarity metrics (e.g., \textit{`Which specific subtype/histological variant of this lesion is present in the images?'}).
Pathologists were instructed to omit the lesion name whenever possible to encourage vision-language models (VLMs) to base their answers solely on the images and to preserve the benchmark’s integrity.
Questions linked to the diagnosis (e.g., prognosis) are therefore potentially more difficult, as the diagnosis is not provided.

The $Q_{\text{mc}}$ questions were grouped into five domains: morphology, ancillary testing, diagnosis, clinical, and mix.
Morphology questions assessed recognition, description, and interpretation of histological features, including growth patterns, tumor differentiation, grade, stage, and cell type of origin.
Ancillary testing questions addressed appropriate confirmatory tests for diagnosis, expected staining patterns, marker expression, and characteristic genetic aberrations.
Diagnosis questions focused on interpreting histological findings and test results, defining differential diagnoses, knowledge of diagnostic criteria, and recognizing specific variants or subtypes.
Clinical questions evaluated background clinical knowledge, including treatment, risk factors, complications, etiology, prognosis, and typical lesion localization.
The mix category combined elements from multiple domains.
$Q_{\text{open}}$ questions followed the same domains, except for the mixed category.

No standardized clinical context was provided; only the tissue type was given, except for $Q_{\text{tissue}}$.
Contributing pathologists could incorporate clinical information within $Q_{\text{mc}}$ or $Q_{\text{open}}$ if desired.

\subsubsection*{Images}

For each question, pathologists annotated up to four rectangular regions of interest (ROIs), ideally at varying magnifications to simulate a pathologist's workflow (see \cref{fig:workflow}).
ROIs could be reused across questions and could originate from one or multiple H\&E- or PAS-stained whole-slide images (WSIs) from the same patient.
For every question, we also provide a low-resolution overview image for each WSI contributing ROIs.
Overviews were automatically extracted from the full WSI; if the pathologist wished to exclude specific areas (e.g., secondary lesions), a manual annotation defined the overview region instead.
Annotations were created in QuPath \cite{Bank17} or HALO \cite{Indi25}, or provided as coordinates.
Each annotated or overview region was extracted as a PNG image at the highest magnification possible, with a maximum size of 1000×1000 pixels or equivalent.
Resolution was 0.10–34 µm/px for ROIs and 4.0–124 µm/px for overviews.

\subsubsection*{Data curation}

As a final data curation step, a board-certified pathologist (F.Me.) reviewed all cases to refine both content and answer options.
During review, the pathologist noted that some contributing pathologists had provided only a single organ or tissue as correct answer for $Q_{\text{tissue}}$, even though multiple tissues could be considered correct, for instance, when other tissues were present and technically correct even if not the primary target (e.g., fatty tissue in a breast resection).
Additional tissue types were added as correct answers when appropriate.
Case-specific questions were excluded or revised if ambiguously phrased, open to multiple interpretations, unclear, or outdated (i.e., inconsistent with current guidelines).

All questions were reviewed to ensure they could be answered using the provided images alone.
Contributing pathologists had access to the full cases and could underestimate the information needed for standalone ROIs.
Because ROIs were defined before PNG image extraction, pathologists could not always anticipate the resulting sharpness and level of detail.
Additional ROIs were added when higher magnification was required to resolve cellular detail or when only high-magnification ROIs were available and broader context was needed.

For $Q_{\text{diagnosis}}$, synonym sets were compiled for each ground truth diagnosis using the WHO Classification of Tumors and ICD-O codes for neoplastic lesions, and ICD-11 codes for all lesions.
Synonyms included common abbreviations, the fully written diagnosis with the abbreviation in parentheses, and UK and US spelling variants.
For example, \textit{benign prostatic hyperplasia} was mapped to \textit{benign prostatic hyperplasia; BPH; benign prostatic hyperplasia (BPH)}.

\subsection*{Reader study}

\subsubsection*{Case selection}

For our reader study, we selected 115 cases (DALPHIN$_{reader}$) from the full dataset of 300 cases (DALPHIN$_{full}$), aiming for a balanced representation across institutions and subspecialties, while including the widest possible range of diagnoses.
The study comprised 95 out of 130 diagnoses, including 33 non-neoplastic entities.

\subsubsection*{Readers}

We defined three levels of expertise: (1) subspecialty \emph{experts}: pathologists with >10 years of experience in the subspecialty and peer recognition (serving as a reference for internal or external second opinions or contributing to guidelines); (2) subspecialty \emph{semi-experts}: pathologists who routinely practice in the subspecialty but are not yet considered experts (<10 years of experience and/or lacking peer recognition); and (3) subspecialty \emph{non-experts}: pathology residents or board-certified pathologists who do not routinely practice in the subspecialty and, to avoid outdated knowledge, have completed residency within the past five years.
For the hepatopancreatobiliary subspecialty, cases were distributed among liver, pancreas, and gastrointestinal experts (the latter for gallbladder cases), rather than being assigned to a single subspecialty expert.
Pathologists could be experts or semi-experts for multiple subspecialties if they met the criteria.
A total of 24 pathologists and seven pathology residents participated in the study.

\subsubsection*{Design}

We conducted the reader study on the Grand Challenge platform \cite{Meak25} (Extended Data \cref{fig:reader-study-interface}).
Each case was reviewed by one expert and three non-experts, and, in seven of the 14 subspecialties (60 cases), also by two semi-experts.
Cases were randomly assigned to non-experts; experts and semi-experts reviewed predetermined sets according to their subspecialties.
Each reader reviewed between three and 55 cases.

Within each reader’s set, cases were presented in random order.
Questions were presented sequentially rather than randomly to better reflect a typical pathologist case review: first $Q_{\text{tissue}}$ (if applicable), followed by $Q_{\text{neoplasm}}$, $Q_{\text{diagnosis}}$, and $Q_{\text{behavior}}$ (if applicable), then $Q_{\text{mc}}$, and, when present, $Q_{\text{open}}$.
$Q_{\text{behavior}}$ was asked only for neoplastic cases, though readers were not informed of this criterion.
To prevent readers from inferring neoplastic status from the presence of $Q_{\text{behavior}}$, $Q_{\text{diagnosis}}$ was always presented first.
Although $Q_{\text{behavior}}$ could have been included for all cases to conceal this pattern, it was limited to relevant cases to reduce reader burden.

A brief preamble was added before each question.
For $Q_{\text{tissue}}$, it read, \textit{`These are images of a histological slide containing a pathological lesion'}, with `\textit{These are images'} changed to \textit{`This is an image'} when only a single image was provided.
For subsequent questions, the tissue type was specified (e.g., \textit{`These are images of a histological slide of lung tissue containing a pathological lesion'}).
After each multiple-choice question, the instruction \textit{`Only indicate the letter of your answer'} was included.

Readers entered answers in a dedicated field, with optional comments, and indicated use of external resources.
Textbooks or journal articles were permitted, but AI tools, including VLMs, were not.
Multiple-choice questions with more than one correct option required selecting a single answer.
Questions had to be answered sequentially, with no backtracking, though the study could be paused and resumed from the first unanswered question; no time limit was set in the study.

\subsubsection*{Response curation}

Some readers provided multiple answers to $Q_{\text{tissue}}$, $Q_{\text{diagnosis}}$, and $Q_{\text{mc}}$; in these cases, the first response was recorded.
An exception was $Q_{\text{tissue}}$ when multiple soft tissue types were listed (e.g., adipose, muscle, fibrous tissue), which was summarized as `soft tissue'.

For $Q_{\text{neoplasm}}$ and $Q_{\text{behavior}}$, some non-expert responses were `unsure' or similar and did not fall into predefined categories ($Q_{\text{neoplasm}}$: yes/no; $Q_{\text{behavior}}$: benign/malignant/in situ/uncertain).
We assumed that respondents would have guessed if forced to respond and randomly assigned a response category using probabilities matching the empirical distribution of categories selected by non-experts, an approach we call \emph{probabilistic imputation}.
Two experts responded to $Q_{\text{behavior}}$ with `premalignant', which is not an ICD-O-3 behavior category.
Because premalignant lesions may be ICD-O-3 coded as benign (e.g., traditional serrated adenoma), in situ, or uncertain behavior, we randomly assigned one of these categories using probabilities matching the empirical distribution of expert responses.

To assess the impact of these assumptions, we recomputed metrics under two alternative scenarios: a `best-case' scenario, in which all responses were considered correct (i.e., modified to match the reference answer), and a `worst-case' scenario, in which all responses were considered incorrect (i.e., modified not to match the reference answer).
For $Q_{\text{behavior}}$, in the worst-case scenario, the two non-expert `unsure' responses and the two expert `premalignant' responses were reassigned to incorrect valid categories, with one assigned to benign and one to malignant for each expertise level.

\subsection*{VLM evaluation}

We compared the pathology-specific VLM PathChat+ \cite{Chen25} to the general-purpose state-of-the-art VLMs GPT-5 \cite{Open25} and Gemini 2.5 Pro \cite{Goog25} using our benchmark.
PathChat+ was evaluated by its development team using its Application Programming Interface (API).
Evaluation of the general-purpose VLMs was performed via the OpenAI Python API bindings (openai, version 1.77.0), with an alternative endpoint being used for Gemini 2.5 Pro (\texttt{gemini-2.5-pro}): \url{https://generativelanguage.googleapis.com/v1beta/openai/}. For both models, reasoning effort was set to `medium' in the OpenAI API request, which translates to a thinking budget of 8,192 tokens for Gemini 2.5 Pro. Images were encoded using the base64 scheme before being added to the request. 
All API calls for both models were made in September 2025.

\subsubsection*{Answer generation scenarios}

We evaluated VLMs on the benchmark under two scenarios: the \emph{independent} and the \emph{contextual} generation scenario.

In the independent generation scenario, each question was treated as a standalone query: the model received the question text and associated image(s) and generated an answer without retaining any context from previous questions.
Incorrect answers did not influence subsequent questions for the same case, preventing error propagation.

In the contextual (feedback) generation scenario, the model processed all questions for a single case sequentially, maintaining a conversation history of prior questions, images, and its own answers to inform subsequent responses.
Questions followed the same order as in the reader study: $Q_{\text{tissue}}$ (if applicable), $Q_{\text{neoplasm}}$, $Q_{\text{diagnosis}}$, $Q_{\text{behavior}}$, $Q_{\text{mc}}$, and finally $Q_{\text{open}}$ (when present).
History was reset between cases, and no corrections were applied to previous answers.
The first question of each case included a special instruction to prime the model: \textit{`The following question(s) and image(s) are related to the same medical case'}.
This mode simulates a more realistic diagnostic workflow, where a pathologist may ask follow-up questions about the same case.
For general-purpose VLMs, images provided for a given question were kept in the conversation history, even if they duplicated images from earlier questions.
For PathChat+, when including all images for the current question, memory constraints required pruning previous images from the conversation history if they were also provided for the current question.
All textual information in the conversation history remained unchanged.

Unlike in the reader study, where $Q_{\text{behavior}}$ was asked only for neoplastic cases, we included it for all cases in the benchmark, although only neoplastic responses were evaluated.
This prevents the presence of $Q_{\text{behavior}}$ from revealing neoplastic status and preserves benchmark integrity.
For non-neoplastic cases, the same images used for $Q_{\text{diagnosis}}$ were provided for $Q_{\text{behavior}}$.

\subsubsection*{Preamble}

A preamble was added before each question, consisting of several components:
(1) Tissue description: For $Q_{\text{tissue}}$, \textit{`These are images of a histological slide containing a pathological lesion'}, with \textit{`These are images'} replaced by \textit{`This is an image'} when only a single image was provided.
For other questions, the tissue type was specified (e.g., \textit{`These are images of a histological slide of lung tissue containing a pathological lesion'}).
(2) Image clarification: Indicated which images were low-resolution overviews and which were ROIs, as some questions specifically targeted ROIs (e.g., \textit{`Is the finding depicted in the regions of interest a feature of high-grade tumors? Briefly explain'}).
Sentences such as \textit{`The first image is a low-resolution overview of a histological slide, and the other images are higher-resolution regions of interest'} were adapted to the number of images.
(3) Lesion type reminder: \textit{`The depicted pathological lesion can be non-neoplastic or neoplastic'}.
This encouraged models to consider both neoplastic and non-neoplastic possibilities, based on pilot study observations \cite{Lems25} that omitting it led models to focus primarily on tumor diagnoses.
(4) Theoretical question disclaimer: \textit{`The following question is just theoretical, no clinical consequence will be drawn based on the answer to this question'}.
This circumvented VLM safety guardrails that would otherwise prevent medical image interpretation.
(5) Answer length instruction: \textit{`Please only give one short answer'}, to prevent lengthy responses.
(6) Multiple-choice instruction: For $Q_{\text{mc}}$ questions, \textit{`Only indicate the letter of your answer'} was appended after the question.

\subsubsection*{Response curation}

Response curation was designed to be consistent and reproducible without requiring expert pathology knowledge, as the public benchmark relies on researcher-submitted VLM responses.
For $Q_{\text{tissue}}$, VLM responses were preprocessed by removing parenthetical content (e.g., \textit{respiratory (bronchial) epithelium} $\rightarrow$ \textit{respiratory epithelium}).
Each cleaned response was matched against our hierarchical tissue taxonomy (see `Evaluation' in Methods) using synonym patterns, with matches ranked by length (longest preferred) and then text position (earliest preferred); the highest-ranked match was retained.
Responses that returned no match (e.g., \textit{`No residual normal tissue is recognizable'}) were retained as-is.
For $Q_{\text{neoplasm}}$ and $Q_{\text{behavior}}$, all VLMs produced consistent, directly evaluable responses.

For $Q_{\text{diagnosis}}$, outputs varied, sometimes including the organ (e.g., \textit{fibroadenoma of the breast}), grade (e.g., \textit{meningioma, CNS Grade 1}), bracketed details (e.g., \textit{myocardial infarction (healed or organizing stage)}), and multiple diagnoses or alternative formulations of the same diagnosis.
The predicted diagnosis was extracted from each model response using standardized rules.
Text in brackets was removed unless it represented an abbreviation (e.g., \textit{inflammatory myofibroblastic tumor (IMT)}) or a WHO classification, in which case the WHO-designated diagnosis was extracted (e.g., \textit{prostatic adenocarcinoma (WHO Classification: acinar adenocarcinoma) $\rightarrow$ acinar adenocarcinoma}), provided it specified a diagnosis rather than a diagnostic group (e.g., \textit{papillary fibroelastoma (benign cardiac tumor, WHO classification) $\rightarrow$ papillary fibroelastoma}).
Grading information was removed unless explicitly stated as low- or high-grade in the model response (e.g., \textit{tubular adenoma with high-grade dysplasia}).
Extraneous descriptive text outside brackets was removed (e.g., \textit{diffuse large B‑cell lymphoma (DLBCL), NOS – hepatic involvement $\rightarrow$ diffuse large B‑cell lymphoma (DLBCL), NOS}).
When multiple diagnoses were present, the first was selected unless a WHO-classified diagnosis was specified.
Organ information formulated as `of the [organ]' and (sub)type/variant information provided after the diagnosis were also removed (e.g., \textit{Hodgkin lymphoma, nodular sclerosis subtype $\rightarrow$ Hodgkin lymphoma}).

For $Q_{\text{mc}}$, if multiple options were returned despite instructions, the first selected option was recorded.
For $Q_{\text{open}}$, the diversity of questions and the subjective nature of a `core answer' prevented consistent, reproducible curation without pathology expertise; therefore, responses were left unchanged.

\subsection*{Evaluation}

Responses to $Q_{\text{tissue}}$ were evaluated using a hierarchical tissue taxonomy adapted from \cite{Moon26}, organizing organs and tissue types with synonyms (e.g., \textit{adrenal cortex}, \textit{cortical adrenal tissue}).
Each response was compared against one or more ground truth terms, and a hierarchical scoring method was applied: 1.0 for an exact match, 0.75 if one step removed (parent, child, or sibling), 0.5 if two steps removed, and 0.0 otherwise (including responses not found in the taxonomy, such as `No residual normal tissue is recognizable').
For multiple ground truth terms, the highest score was used.

For $Q_{\text{neoplasm}}$, $Q_{\text{behavior}}$, and $Q_{\text{mc}}$, correctness was assessed against the ground truth.
Metrics for $Q_{\text{neoplasm}}$ included binary classification metrics (F1 score, Matthews Correlation Coefficient [MCC], precision, recall, specificity); for $Q_{\text{behavior}}$, multiclass classification metrics (Cohen's $\kappa$, MCC) and class-specific F1 and MCC scores for benign and malignant; and for $Q_{\text{mc}}$, accuracy.
Multiple-choice questions with several correct options were considered correct if any matched the response.
$Q_{\text{mc}}$ responses that did not indicate an answer option (e.g., `Not applicable') were considered incorrect.

$Q_{\text{diagnosis}}$ and $Q_{\text{open}}$ were evaluated using traditional and semantic natural language processing metrics.
All responses and ground truth answers were converted to lowercase.
For $Q_{\text{diagnosis}}$, when multiple ground truth terms (diagnosis synonyms) existed, responses were evaluated against all terms and the highest score retained.
Traditional metrics included BLEU-1, ROUGE-L, CIDEr, and METEOR; BLEU-4 was omitted because ground truth answers were typically too concise.
Semantic evaluation used BERT-based approaches \cite{Zhan20,Lee20}, including BERTScore (precision, recall, F1) and BioBERT similarity to capture meaning-level agreement.
For BioBERT, we used the pretrained model at \url{https://huggingface.co/pritamdeka/BioBERT-mnli-snli-scinli-scitail-mednli-stsb}.

\subsection*{Statistics}

For expertise levels with multiple pathologist responses per question (semi-experts and non-experts), we calculated metrics by randomly sampling one pathologist per question, computing the metric, and repeating this $n = 1{,}000$ times to obtain the mean.
We applied the same resampling strategy for nonparametric bootstrapping ($n = 1{,}000$ replicates) over the set of questions to estimate 95\% confidence intervals.
We computed relative performance differences as percentage change relative to the reference group
\((\text{comparison} - \text{reference}) / \text{reference} \times 100\), with positive values indicating higher performance.

We tested observed differences between pairs of VLMs or expertise levels within a dataset subset (e.g., DALPHIN$_{reader}$) for statistical significance using a two-sided paired permutation test ($n = 1{,}000$ permutations), with the null hypothesis of no difference.
In each permutation, we randomly swapped answers between models or expertise levels to obtain a new difference, applying the same procedure of sampling one pathologist per question for semi-experts and non-experts. 
The $P$-value was the proportion of permuted differences whose absolute value exceeded the observed difference.
We made no adjustments for multiple testing.
We treated subgroup analyses (e.g., by subspecialty or neoplastic status) as exploratory and descriptive and performed no statistical testing.

\subsection*{Ethics statement}

The use of cases for this study was approved by the institutional review boards of RUMC (reference 2024-17419), AUMC (reference AJOUIRB-DB-2025-033), and Biopticka (reference 240611).
For VPC, IMP Diagnostics, and HUGTiP, ethical approval was not required because the study was retrospective and the data were fully anonymized, in accordance with local and national regulations.

\section*{Data Availability}

The DALPHIN benchmark images and questions are available on Zenodo (\url{https://zenodo.org/records/18609450}) under a CC BY-NC-ND 4.0 license.
Reference answers are sequestered and only used for automatic performance evaluation through \url{https://dalphin.grand-challenge.org/}, ensuring robust and reproducible evaluation of pathology AI copilots.

\section*{Code Availability}

The code is available in our GitHub repository (\url{https://github.com/computationalpathologygroup/DALPHIN}).
It includes scripts to download the dataset from Zenodo, a reference implementation for running a VLM on DALPHIN, and evaluation code identical to that used on Grand Challenge for scoring model submissions.

\bibliography{sample}

\section*{Acknowledgements}

The authors thank Milda Pocevičiūtė for her contributions to the pilot study underlying this work and for her support in obtaining data from VPC.
The authors also thank Alon Vigdorovits for his advice on reader study design and contributions to the evaluation pipeline.

This project was funded by the Dutch Cancer Society (KWF, COMMITMENT project number 15386) and the Dutch Research Council (NWO AES Domain, VIDI grant number 18388), and was supported by the Ammodo Science Award 2024 and a fellowship from Stichting Hanarth Fonds (the Netherlands).

\section*{Author Contributions}

C.L. and F.C. conceived the study.
C.L., S.M., N.Kh., F.Me., and F.C. designed the experiments.
N.Ku., B.B., T.L., S.K., V.V., L.P., S.H., M.E.S.F., P.L.F., J.D., D.P., R.A., A.L., D.O., D.M., A.B.B., D.v.M., A.M.V., S.V., J.v.I., and M.B. collected and provided the data.
C.L. and F.Me. processed and curated the data.
C.L. conducted the reader study and collected the responses.
N.Ku., J.D., D.P., D.O., D.v.M., A.M.V., J.v.I., K.W., I.N., K.H., U.F., K.G., J.S., B.S.C., J.T.S., E.M., L.C., G.Q., Y.G.B., J.W.F., G.J.L.H.v.L., J.H.v.d.T., L.A.A.B., R.R.d.K., P.W., S.F., M.M., A.K., R.B., D.T., and J.L.P. participated in the reader study.
S.M. ran the general-purpose VLMs on the benchmark.
M.Y.L., C.C., and F.Ma. ran PathChat+ on the benchmark.
C.L. and S.M. performed the experimental analysis, including model performance evaluation, and interpreted the results.
S.M., N.Kh., F.Me., and F.C. provided feedback on the analysis.
R.D., J.S.F.M., and J.v.d.L. provided feedback on the reader study design and statistical analysis plan.
F.C. supervised the work.
C.L., S.M., N.Kh., F.Me., and F.C. wrote the main manuscript.
All authors reviewed the manuscript and approved its contents.

\section*{Competing Interests}

B.B. and T.L. are employed by Lunit, an AI medical product company.
M.B. is medical advisor at Aiosyn BV, the Netherlands.
J.T.S. received research support from Palex Medical and honoraria or consultation fees from CellsIA, MSD, Roche, and 3DHistech.
J.v.d.L. was a member of the advisory boards of Philips, the Netherlands, and ContextVision, Sweden, and received research funding from Philips, the Netherlands, ContextVision, Sweden, and Sectra, Sweden, in the last five years. He is Chief Scientific Officer (CSO) and shareholder of Aiosyn BV, the Netherlands.
F.C. was chair of the Scientific and Medical Advisory Board of TRIBVN Healthcare, France, and received advisory board fees from TRIBVN Healthcare, France, in the last five years. He is a shareholder of Aiosyn BV, the Netherlands.
All other authors declare no conflict of interest.

\nolinenumbers
\section*{Extended Data}

\captionsetup[figure]{name=Extended Data Figure}
\setcounter{figure}{0}

\begin{figure}[ht]
\centering
\includegraphics[width=1.0\textwidth]{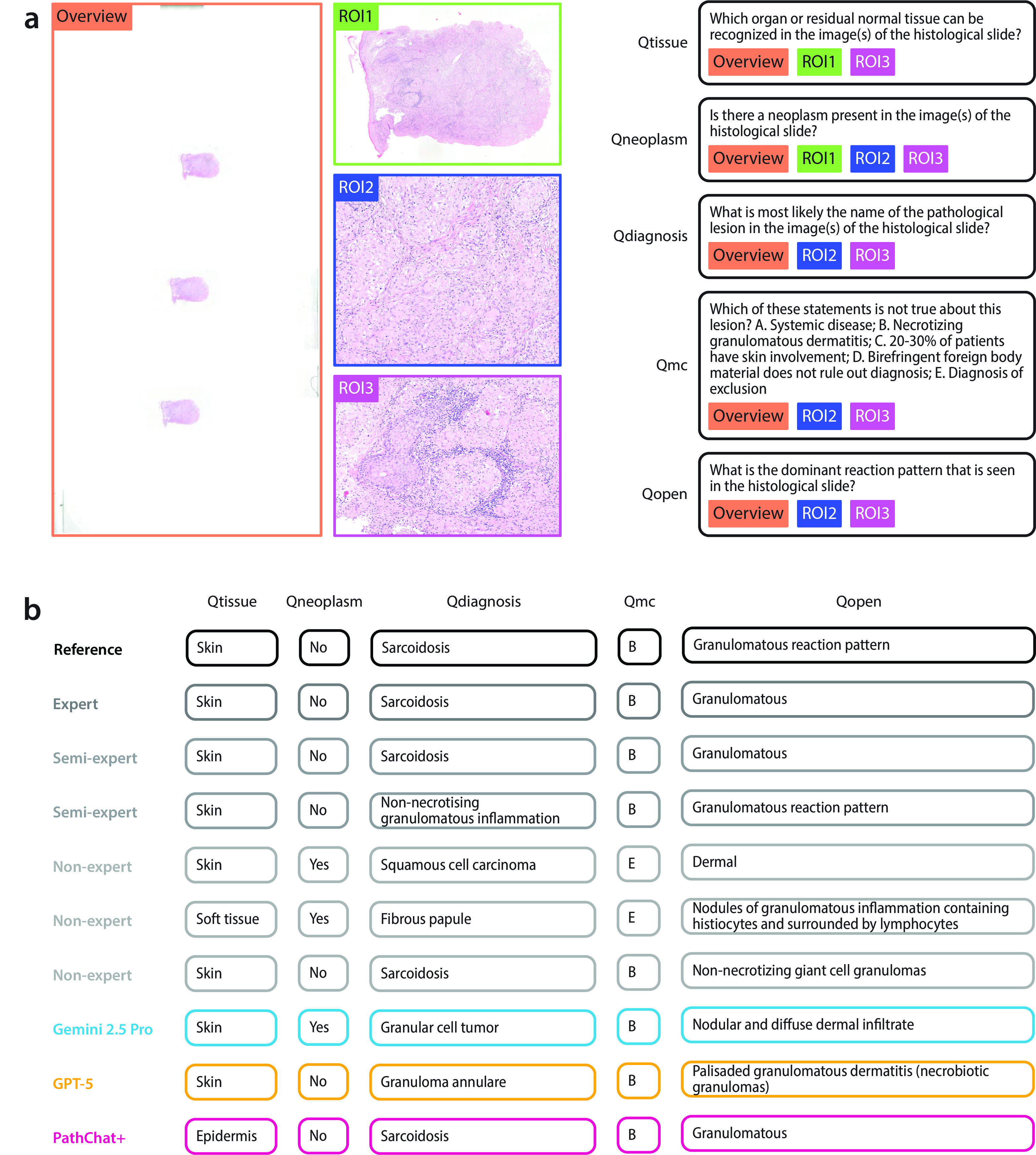}
\caption{
(a) Example DALPHIN case, including images and associated questions.
(b) Responses from subspecialty expert, semi-expert, and non-expert (resident) pathologists, as well as VLMs, to all questions in the same DALPHIN case, compared against the reference standard (i.e., the answers provided by the pathologist who contributed the case). Note: this example was excluded from the benchmark to maintain its integrity.
}
\label{fig:example-case}
\end{figure}

\begin{figure}[ht]
\centering
\includegraphics[width=0.8\textwidth]{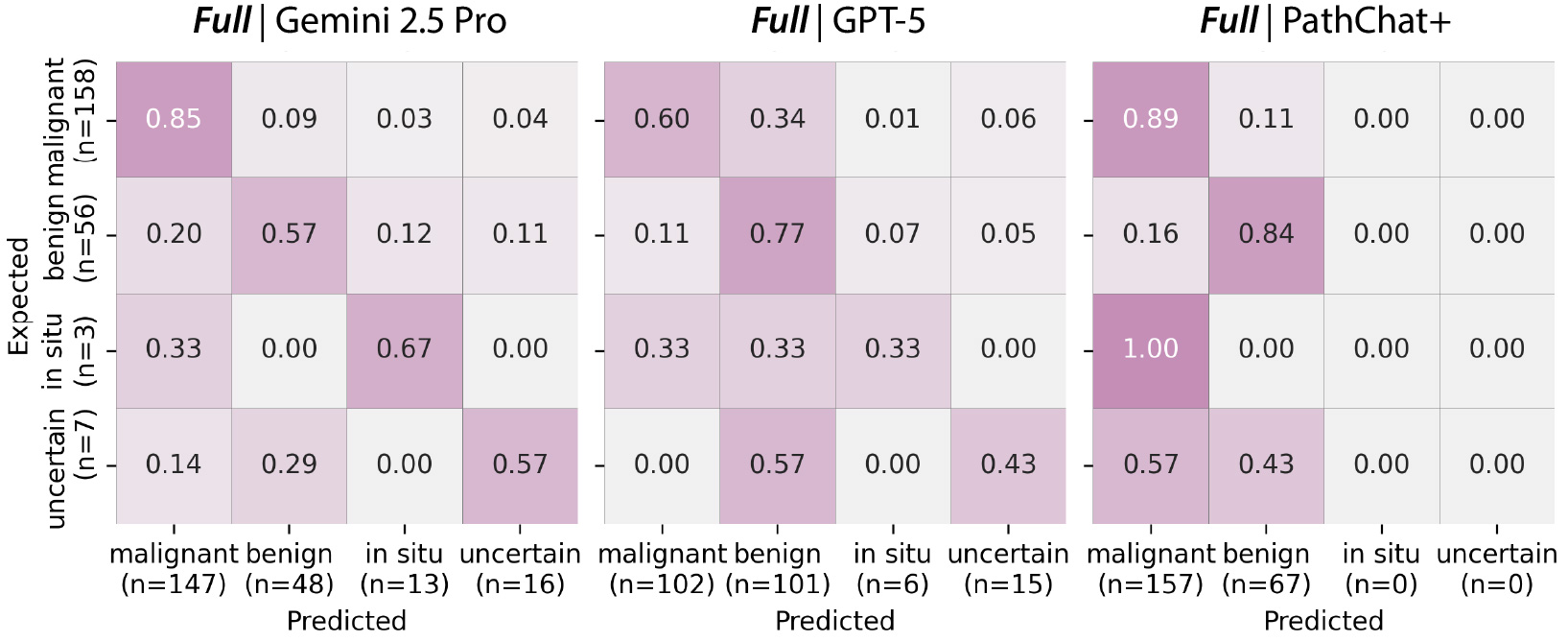}
\caption{Confusion matrices for neoplastic behavior classification ($Q_{\text{behavior}}$) for VLMs in DALPHIN$_{full}$.
}
\label{fig:cm-full}
\end{figure}

\begin{figure}[ht]
\centering
\includegraphics[width=0.7\textwidth]{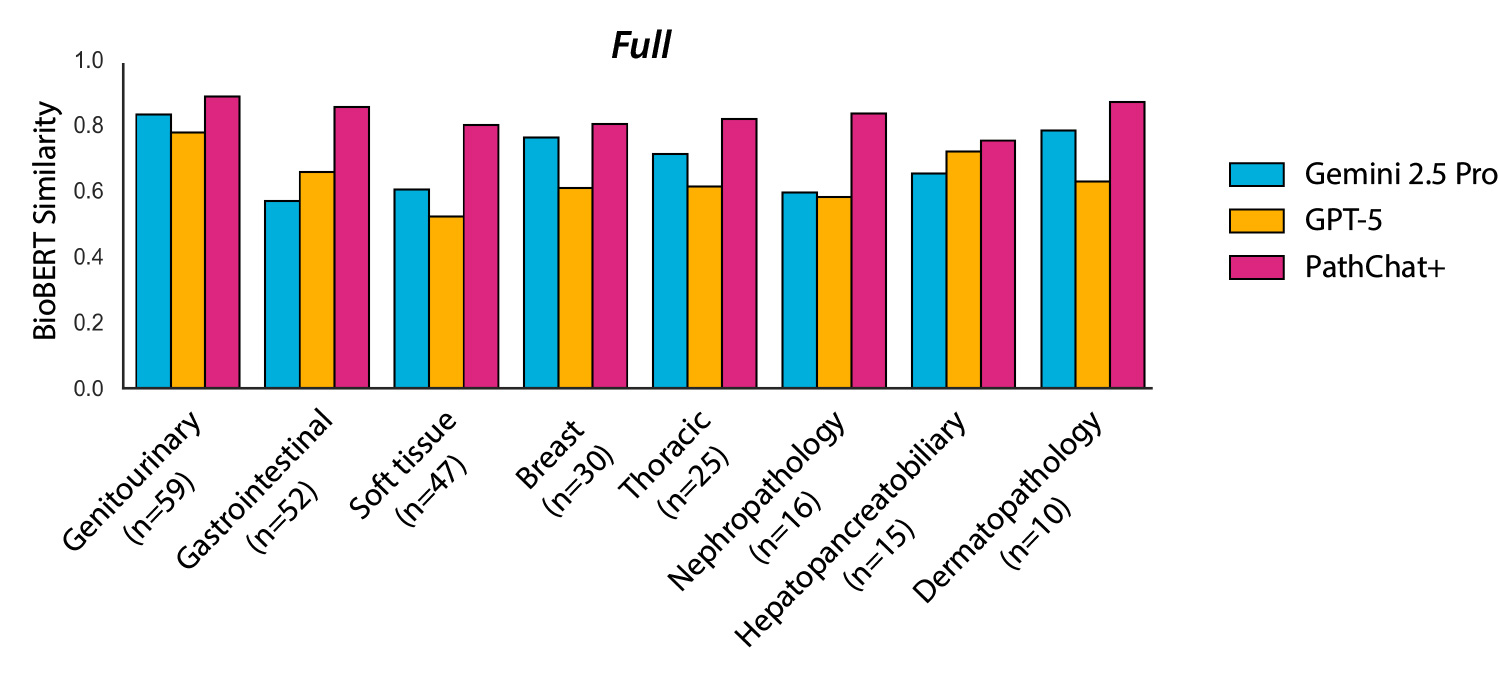}
\caption{BioBERT similarity scores for a free-response diagnosis task ($Q_{\text{diagnosis}}$) for VLMs in DALPHIN$_{full}$, stratified by subspecialty. Only subspecialties with $\geq$10 cases are shown.
}
\label{fig:diagnosis-subspecialty}
\end{figure}

\begin{figure}[ht]
\centering
\includegraphics[width=1.0\textwidth]{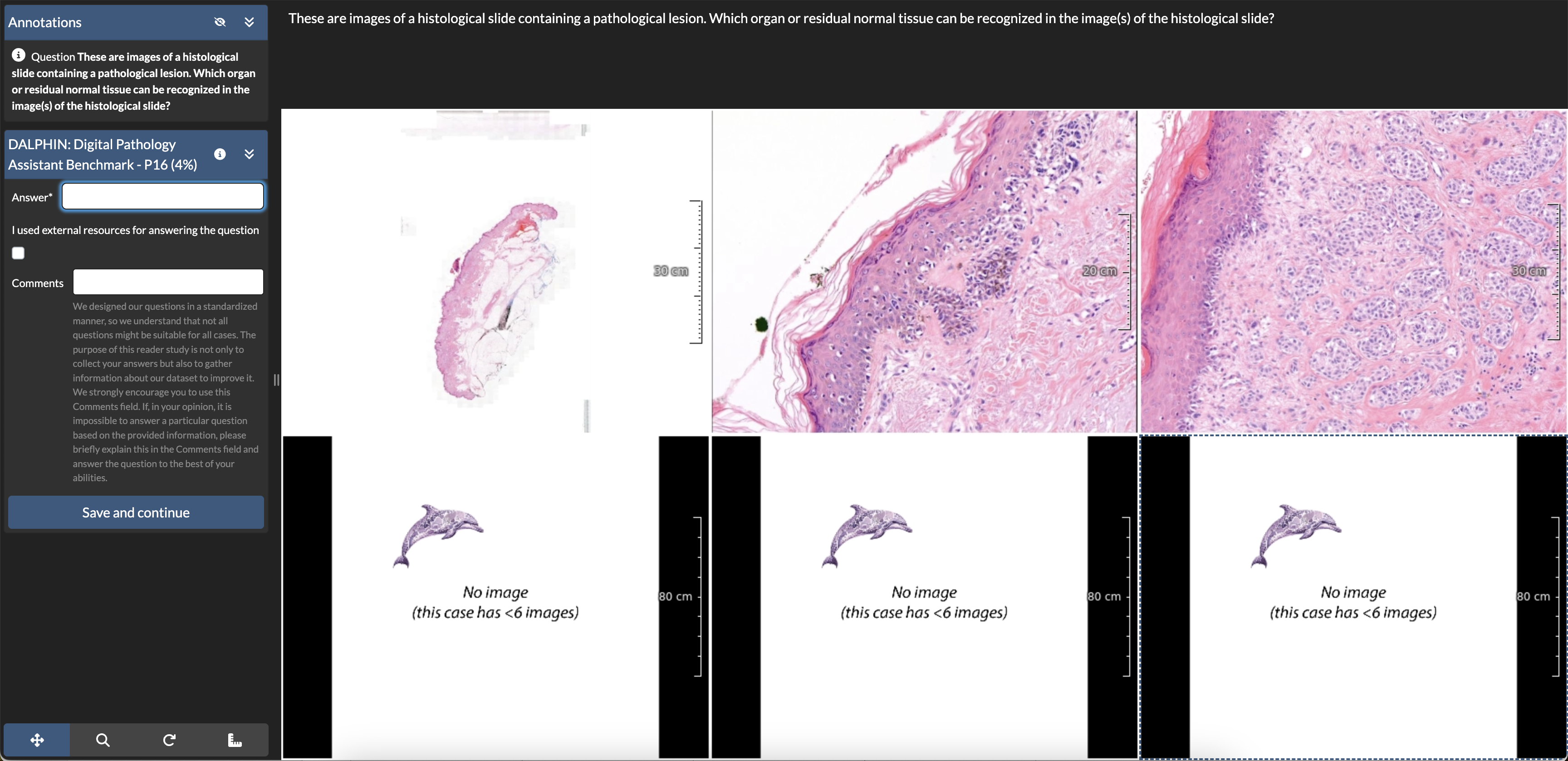}
\caption{Screenshot of the reader study interface on the Grand Challenge platform. Each question was displayed twice: once at the top above the images and again in the top-left corner (under `Annotations'). Readers entered answers in a dedicated field on the left panel, could provide optional comments, and indicated whether external resources were used.
}
\label{fig:reader-study-interface}
\end{figure}

\FloatBarrier

\section*{Supplementary Information}

\begin{table}[htbp]
    \centering
    \caption{\texttt{supplementary\_table\_1.csv} in the supplementary files. Exact $P$-values for all statistical comparisons reported in the Results. Each row corresponds to a comparison between a pair of raters (models or pathology expertise levels) for a given task and data subset. Models are evaluated under independent or contextual answer generation scenarios (\texttt{\{model\}-independent} or \texttt{\{model\}-contextual}) (see `VLM evaluation' in Methods). The table reports the number of cases ($n$), the metric for each rater with 95\% bootstrapped confidence intervals, the mean difference between raters, and the corresponding $P$-value.}
    \label{tab:p_values}
\end{table}

\begin{table}[ht]
    \centering
\begin{tabular}{llll}
    \hline
    Model/Pathology & DALPHIN$_{full}$ & DALPHIN$_{reader}$ & DALPHIN$_{semi}$ \\
    expertise level & ($n=272$) & ($n=101$) & ($n=54$) \\
    \hline
    Gemini 2.5 Pro & 0.734 (0.683–0.781) & 0.792 (0.713–0.866) & 0.875 (0.773–0.954) \\
    GPT-5 & 0.436 (0.378–0.495) & 0.438 (0.349–0.530) & 0.509 (0.380–0.639) \\
    PathChat+ & 0.787 (0.737–0.836) & 0.762 (0.681–0.839) & 0.810 (0.699–0.907) \\
    Experts & NA & 0.948 (0.899–0.985) & 0.963 (0.907–1.000) \\
    Non-experts & NA & 0.782 (0.683–0.876) & 0.770 (0.643–0.889) \\
    Semi-experts & NA & NA & 0.868 (0.759–0.963) \\
    \hline
    \end{tabular}
    \caption{Organ recognition score on a free-response organ recognition task ($Q_{\text{tissue}}$). 95\% confidence intervals from bootstrapping are included in parentheses.}
    \label{tab:q_tissue}
\end{table}

\begin{table}[ht]
    \centering
    \begin{tabular}{lll}
    \hline
    Subspecialty & Model & DALPHIN$_{full}$ \\
    \hline
    Genitourinary ($n=58$) & Gemini 2.5 Pro & 0.836 (0.741–0.927) \\
     & GPT-5 & 0.466 (0.328–0.603) \\
     & PathChat+ & 0.914 (0.828–0.983) \\
    \hline
    Gastrointestinal ($n=50$) & Gemini 2.5 Pro & 0.790 (0.680–0.885) \\
     & GPT-5 & 0.570 (0.440–0.690) \\
     & PathChat+ & 0.855 (0.755–0.935) \\
    \hline
    Soft tissue ($n=41$) & Gemini 2.5 Pro & 0.433 (0.287–0.579) \\
     & GPT-5 & 0.329 (0.195–0.464) \\
     & PathChat+ & 0.409 (0.256–0.567) \\
    \hline
    Breast ($n=29$) & Gemini 2.5 Pro & 0.828 (0.655–0.931) \\
     & GPT-5 & 0.276 (0.103–0.448) \\
     & PathChat+ & 0.931 (0.828–1.000) \\
    \hline
    Thoracic ($n=23$) & Gemini 2.5 Pro & 0.826 (0.652–0.957) \\
     & GPT-5 & 0.522 (0.347–0.739) \\
     & PathChat+ & 0.913 (0.783–1.000) \\
    \hline
    Nephropathology ($n=16$) & Gemini 2.5 Pro & 1.000 (1.000–1.000) \\
     & GPT-5 & 0.312 (0.125–0.562) \\
     & PathChat+ & 0.938 (0.812–1.000) \\
    \hline
    Hepatopancreatobiliary ($n=14$) & Gemini 2.5 Pro & 0.643 (0.357–0.857) \\
     & GPT-5 & 0.429 (0.143–0.714) \\
     & PathChat+ & 0.929 (0.786–1.000) \\
    \hline
    Dermatopathology ($n=10$) & Gemini 2.5 Pro & 0.800 (0.500–1.000) \\
     & GPT-5 & 0.800 (0.500–1.000) \\
     & PathChat+ & 0.900 (0.700–1.000) \\
    \hline
    \end{tabular}
    \caption{Organ recognition score on a free-response organ recognition task ($Q_{\text{tissue}}$), stratified by subspecialty. Only subspecialties with $\geq$10 cases are shown. 95\% confidence intervals from bootstrapping are included in parentheses.}
    \label{tab:q_tissue_subspecialty}
\end{table}

\begin{table}[ht]
    \centering
    \begin{tabular}{lllll}
    \hline
    Metric & Model/Pathology & DALPHIN$_{full}$ & DALPHIN$_{reader}$ & DALPHIN$_{semi}$ \\
     & expertise level & ($n=300$) & ($n=115$) & ($n=60$) \\
    \hline
    F1 & Gemini 2.5 Pro & 0.925 (0.901–0.947) & 0.891 (0.840–0.939) & 0.917 (0.857–0.970) \\
     & GPT-5 & 0.844 (0.803–0.879) & 0.809 (0.733–0.878) & 0.785 (0.676–0.874) \\
     & PathChat+ & 0.968 (0.951–0.982) & 0.945 (0.906–0.978) & 0.967 (0.923–1.000) \\
     & Experts & NA & 0.969 (0.936–0.993) & 0.978 (0.938–1.000) \\
     & Non-experts & NA & 0.928 (0.875-0.972) & 0.928 (0.851-0.988) \\
     & Semi-experts & NA & NA & 0.966 (0.905–1.000) \\
\hline
    \end{tabular}
    \caption{Binary classification metrics for neoplastic status (Q$_{neoplasm}$). 95\% confidence intervals from bootstrapping are included in parentheses.}
    \label{tab:q_neoplasm}
\end{table}

\begin{table}[ht]
    \centering
    \begin{tabular}{llll}
    \hline
    Metric& Response handling & DALPHIN$_{reader}$ & DALPHIN$_{semi}$ \\
     & strategy & ($n=115$) & ($n=60$) \\
    \hline
    F1 & Probability-based & 0.928 (0.875-0.972) & 0.928 (0.851-0.988) \\
     & Worst-case & 0.922 (0.863-0.967) & 0.925 (0.845-0.980) \\
     & Best-case & 0.936 (0.883-0.978) & 0.931 (0.857-0.989) \\
    \hline
    MCC & Probability-based & 0.760 (0.592-0.905) & 0.699 (0.392-0.934) \\
     & Worst-case & 0.739 (0.562-0.888) & 0.686 (0.373-0.918) \\
     & Best-case & 0.787 (0.627-0.928) & 0.714 (0.414-0.944) \\
    \hline
    Precision & Probability-based & 0.904 (0.822-0.973) & 0.910 (0.796-1.000) \\
     & Worst-case & 0.899 (0.816-0.973) & 0.910 (0.795-1.000) \\
     & Best-case & 0.915 (0.840-0.977) & 0.916 (0.809-1.000) \\
    \hline
    Recall & Probability-based & 0.953 (0.886-1.000) & 0.949 (0.867-1.000) \\
     & Worst-case & 0.945 (0.873-1.000) & 0.941 (0.844-1.000) \\
     & Best-case & 0.958 (0.899-1.000) & 0.949 (0.867-1.000) \\
    \hline
    Specificity & Probability-based & 0.777 (0.611-0.944) & 0.711 (0.400-1.000) \\
     & Worst-case & 0.768 (0.583-0.917) & 0.711 (0.400-1.000) \\
     & Best-case & 0.805 (0.639-0.944) & 0.734 (0.465-1.000) \\
    \hline
    \end{tabular}
    \caption{Binary classification metrics for neoplastic status (Q$_{neoplasm}$) for non-expert (resident) pathologists under different strategies for handling inconclusive responses. Strategies include probabilistic imputation, worst-case (forcing the response not to match the reference), and best-case (forcing the response to match the reference) (see `Reader study' in Methods). 95\% confidence intervals from bootstrapping are included in parentheses.}
    \label{tab:q_neoplasm_imputation}
\end{table}

\begin{table}[ht]
    \centering
    \begin{tabular}{llll}
    \hline
    Metric & Model/Pathology & DALPHIN$_{full}$ & DALPHIN$_{reader}$ \\
     & expertise level & ($n$=224) & ($n$=78) \\
    \hline
    Cohen's $\kappa$ & Gemini 2.5 Pro & 0.517 (0.414–0.614) & 0.620 (0.470–0.773) \\
     & GPT-5 & 0.351 (0.251–0.441) & 0.377 (0.242–0.520) \\
     & PathChat+ & 0.627 (0.520–0.723) & 0.708 (0.526–0.854) \\
     & Experts & NA & 0.775 (0.630-0.894) \\
     & Non-experts & NA & 0.645 (0.455-0.821) \\
    \hline
    MCC & Gemini 2.5 Pro & 0.523 (0.421–0.622) & 0.631 (0.491–0.778) \\
\hline
    \end{tabular}
    \caption{Multiclass classification metrics for neoplastic behavior (Q$_{behavior}$). 95\% confidence intervals from bootstrapping are included in parentheses.}
    \label{tab:q_behavior}
\end{table}

\begin{table}[ht]
    \centering
    \begin{tabular}{llll}
    \hline
    Metric & Pathology & Response handling & DALPHIN$_{reader}$ \\
     & expertise level & strategy & ($n=115$) \\
    \hline
    Cohen's $\kappa$ & Experts & Probability-based & 0.775 (0.630-0.894) \\
     &  & Worst-case & 0.772 (0.623-0.893) \\
     &  & Best-case & 0.822 (0.694-0.928) \\
     & Non-experts & Probability-based & 0.645 (0.455-0.821) \\
     &  & Worst-case & 0.644 (0.453-0.821) \\
     &  & Best-case & 0.661 (0.469-0.835) \\
\hline
    \end{tabular}
    \caption{Multiclass classification metrics for neoplastic behavior (Q$_{behavior}$) for expert and non-expert pathologists under different strategies for handling inconclusive responses. Strategies include probabilistic imputation, worst-case (forcing the response not to match the reference), and best-case (forcing the response to match the reference) (see `Reader study' in Methods). 95\% confidence intervals from bootstrapping are included in parentheses.}
    \label{tab:q_behavior_imputation}
\end{table}

\begin{table}[ht]
    \centering
    \begin{tabular}{llllll}
    \hline
    Subset & Model/Pathology & BioBERT sim. & BERTF1 & BERTPrecision & BERTRecall \\
     & expertise level &  &  &  &  \\
    \hline
    DALPHIN$_{full}$ & Gemini 2.5 Pro & 0.71 (0.68–0.74) & 0.83 (0.82–0.85) & 0.83 (0.82–0.85) & 0.84 (0.82–0.85) \\
    ($n=300$) & GPT-5 & 0.66 (0.63–0.69) & 0.79 (0.77–0.81) & 0.79 (0.78–0.81) & 0.80 (0.78–0.82) \\
     & PathChat+ & 0.85 (0.82–0.87) & 0.89 (0.88–0.91) & 0.90 (0.88–0.91) & 0.89 (0.88–0.91) \\
    DALPHIN$_{reader}$ & Gemini 2.5 Pro & 0.71 (0.66–0.76) & 0.84 (0.82–0.87) & 0.84 (0.82–0.87) & 0.85 (0.83–0.88) \\
    ($n=115$) & GPT-5 & 0.66 (0.61–0.71) & 0.79 (0.76–0.82) & 0.79 (0.76–0.81) & 0.80 (0.77–0.83) \\
     & PathChat+ & 0.86 (0.82–0.90) & 0.91 (0.88–0.93) & 0.91 (0.89–0.93) & 0.91 (0.89–0.93) \\
     & Experts & 0.88 (0.84–0.92) & 0.92 (0.89–0.94) & 0.92 (0.89–0.94) & 0.92 (0.90–0.95) \\
     & Non-experts & 0.79 (0.73–0.85) & 0.86 (0.82–0.90) & 0.86 (0.83–0.90) & 0.87 (0.83–0.90) \\
    DALPHIN$_{semi}$ & Gemini 2.5 Pro & 0.74 (0.67–0.80) & 0.85 (0.82–0.89) & 0.85 (0.82–0.89) & 0.86 (0.83–0.90) \\
    ($n=60$) & GPT-5 & 0.64 (0.58–0.72) & 0.79 (0.76–0.83) & 0.79 (0.75–0.83) & 0.80 (0.77–0.84) \\
     & PathChat+ & 0.85 (0.78–0.90) & 0.90 (0.86–0.93) & 0.90 (0.87–0.93) & 0.90 (0.87–0.93) \\
     & Experts & 0.87 (0.82–0.92) & 0.91 (0.88–0.95) & 0.91 (0.87–0.94) & 0.92 (0.89–0.95) \\
     & Non-experts & 0.78 (0.69–0.86) & 0.86 (0.81–0.91) & 0.86 (0.81–0.91) & 0.87 (0.82–0.92) \\
     & Semi-experts & 0.84 (0.77–0.91) & 0.89 (0.84–0.94) & 0.89 (0.85–0.94) & 0.89 (0.84–0.94) \\
    \hline
    \end{tabular}
    \caption{Semantic NLP metric scores for a free-response diagnosis task ($Q_{\text{diagnosis}}$). 95\% confidence intervals from bootstrapping are included in parentheses.}
    \label{tab:q_diagnosis_sem}
\end{table}

\begin{table}[ht]
    \centering
    \begin{tabular}{llllll}
    \hline
    Subset & Model/Pathology & BLEU-1 & ROUGE-L & METEOR & CIDEr \\
     & expertise level &  &  &  &  \\
    \hline
    DALPHIN$_{full}$ & Gemini 2.5 Pro & 0.49 (0.44–0.54) & 0.50 (0.45–0.55) & 0.44 (0.38–0.49) & 2.67 (2.30–3.05) \\
    ($n=300$) & GPT-5 & 0.38 (0.33–0.42) & 0.41 (0.36–0.46) & 0.33 (0.28–0.37) & 2.16 (1.81–2.50) \\
     & PathChat+ & 0.66 (0.61–0.71) & 0.69 (0.64–0.73) & 0.60 (0.55–0.65) & 3.78 (3.43–4.17) \\
    DALPHIN$_{reader}$ & Gemini 2.5 Pro & 0.50 (0.41–0.58) & 0.51 (0.43–0.60) & 0.46 (0.37–0.54) & 2.58 (2.02–3.10) \\
    ($n=115$) & GPT-5 & 0.37 (0.30–0.44) & 0.40 (0.33–0.48) & 0.32 (0.25–0.39) & 2.01 (1.52–2.51) \\
     & PathChat+ & 0.70 (0.62–0.76) & 0.72 (0.65–0.79) & 0.66 (0.58–0.73) & 3.87 (3.32–4.43) \\
     & Experts & 0.78 (0.71–0.84) & 0.80 (0.73–0.86) & 0.74 (0.67–0.81) & 3.90 (3.39–4.45) \\
     & Non-experts & 0.58 (0.48–0.67) & 0.61 (0.51–0.70) & 0.54 (0.44–0.64) & 3.14 (2.48–3.78) \\
    DALPHIN$_{semi}$ & Gemini 2.5 Pro & 0.53 (0.42–0.64) & 0.55 (0.44–0.65) & 0.46 (0.34–0.58) & 2.62 (1.82–3.42) \\
    ($n=60$) & GPT-5 & 0.38 (0.29–0.49) & 0.42 (0.32–0.53) & 0.33 (0.24–0.43) & 2.03 (1.37–2.79) \\
     & PathChat+ & 0.69 (0.59–0.78) & 0.71 (0.61–0.80) & 0.63 (0.52–0.73) & 3.90 (3.08–4.71) \\
     & Experts & 0.76 (0.67–0.84) & 0.79 (0.70–0.87) & 0.71 (0.61–0.81) & 4.03 (3.27–4.91) \\
     & Non-experts & 0.60 (0.47–0.74) & 0.62 (0.49–0.76) & 0.56 (0.43–0.71) & 3.33 (2.38–4.42) \\
     & Semi-experts & 0.68 (0.57–0.80) & 0.72 (0.61–0.83) & 0.64 (0.52–0.77) & 3.48 (2.71–4.43) \\
    \hline
    \end{tabular}
    \caption{Traditional NLP metric scores for a free-response diagnosis task ($Q_{\text{diagnosis}}$). 95\% confidence intervals from bootstrapping are included in parentheses.}
    \label{tab:q_diagnosis_trad}
\end{table}

\begin{table}[ht]
    \centering
    \begin{tabular}{llllll}
    \hline
    Difficulty & Model & BioBERT sim. & BERTF1 & BERTPrecision & BERTRecall \\
    \hline
    Basic ($n=200$) & Gemini 2.5 Pro & 0.73 (0.69–0.76) & 0.85 (0.83–0.87) & 0.85 (0.82–0.87) & 0.85 (0.83–0.87) \\
    & GPT-5 & 0.69 (0.65–0.72) & 0.81 (0.78–0.83) & 0.81 (0.79–0.83) & 0.81 (0.79–0.84) \\
     & PathChat+ & 0.87 (0.84–0.90) & 0.90 (0.89–0.92) & 0.91 (0.89–0.93) & 0.90 (0.89–0.92) \\
    Broad ($n=46$) & Gemini 2.5 Pro & 0.71 (0.62–0.79) & 0.84 (0.79–0.88) & 0.85 (0.80–0.89) & 0.84 (0.79–0.88) \\
    & GPT-5 & 0.67 (0.59–0.76) & 0.80 (0.76–0.85) & 0.81 (0.76–0.85) & 0.80 (0.76–0.85) \\
     & PathChat+ & 0.84 (0.78–0.91) & 0.88 (0.84–0.92) & 0.89 (0.85–0.93) & 0.88 (0.84–0.92) \\
    In-depth ($n=54$) & Gemini 2.5 Pro & 0.65 (0.58–0.71) & 0.78 (0.74–0.82) & 0.78 (0.74–0.82) & 0.79 (0.75–0.82) \\
     & GPT-5 & 0.57 (0.51–0.63) & 0.73 (0.70–0.77) & 0.73 (0.69–0.76) & 0.75 (0.72–0.78) \\
     & PathChat+ & 0.77 (0.70–0.84) & 0.86 (0.82–0.90) & 0.86 (0.82–0.90) & 0.86 (0.82–0.90) \\
    \hline
    \end{tabular}
    \caption{Semantic NLP metric scores for a free-response diagnosis task ($Q_{\text{diagnosis}}$) for DALPHIN$_{full}$, stratified by diagnostic difficulty (basic [A], broad [B], in-depth knowledge [C], per Dutch pathology education guidelines). 95\% confidence intervals from bootstrapping are included in parentheses.}
    \label{tab:q_diagnosis_difficulty_full_sem}
\end{table}

\begin{table}[ht]
    \centering
    \begin{tabular}{llllll}
    \hline
    Difficulty & Model & BLEU-1 & ROUGE-L & METEOR & CIDEr \\
    \hline
    Basic ($n=200$) & Gemini 2.5 Pro & 0.53 (0.47–0.59) & 0.55 (0.49–0.61) & 0.46 (0.40–0.53) & 3.08 (2.60–3.57) \\
     & GPT-5 & 0.43 (0.37–0.48) & 0.47 (0.41–0.52) & 0.37 (0.31–0.42) & 2.45 (2.01–2.91) \\
     & PathChat+ & 0.70 (0.65–0.75) & 0.73 (0.68–0.78) & 0.63 (0.57–0.68) & 4.15 (3.67–4.59) \\
    Broad ($n=46$) & Gemini 2.5 Pro & 0.53 (0.39–0.67) & 0.54 (0.40–0.68) & 0.50 (0.36–0.64) & 2.40 (1.69–3.22) \\
    & GPT-5 & 0.39 (0.27–0.52) & 0.42 (0.30–0.55) & 0.38 (0.25–0.51) & 2.49 (1.56–3.53) \\
     & PathChat+ & 0.64 (0.51–0.76) & 0.67 (0.55–0.79) & 0.59 (0.46–0.73) & 3.50 (2.62–4.45) \\
    In-depth ($n=54$) & Gemini 2.5 Pro & 0.29 (0.18–0.41) & 0.29 (0.18–0.41) & 0.27 (0.16–0.39) & 1.35 (0.70–2.04) \\
    & GPT-5 & 0.17 (0.08–0.27) & 0.18 (0.09–0.28) & 0.15 (0.07–0.25) & 0.79 (0.28–1.42) \\
     & PathChat+ & 0.53 (0.41–0.65) & 0.54 (0.42–0.66) & 0.51 (0.38–0.64) & 2.65 (1.93–3.39) \\
    \hline
    \end{tabular}
    \caption{Traditional NLP metric scores for a free-response diagnosis task ($Q_{\text{diagnosis}}$) for DALPHIN$_{full}$, stratified by diagnostic difficulty (basic [A], broad [B], in-depth knowledge [C], per Dutch pathology education guidelines). 95\% confidence intervals from bootstrapping are included in parentheses.}
    \label{tab:q_diagnosis_difficulty_full_trad}
\end{table}

\begin{table}[ht]
    \centering
    \begin{tabular}{llllll}
    \hline
    Difficulty & Model/Pathology & BioBERT sim. & BERTF1 & BERTPrecision & BERTRecall \\
     & expertise level &  &  &  &  \\
    \hline
    Basic & Gemini 2.5 Pro & 0.71 (0.65–0.78) & 0.84 (0.81–0.87) & 0.84 (0.81–0.88) & 0.85 (0.82–0.88) \\
    ($n=72$) & GPT-5 & 0.68 (0.62–0.74) & 0.80 (0.77–0.84) & 0.80 (0.77–0.84) & 0.82 (0.78–0.85) \\
     & PathChat+ & 0.88 (0.83–0.92) & 0.91 (0.89–0.94) & 0.92 (0.89–0.94) & 0.92 (0.89–0.94) \\
     & Experts & 0.84 (0.78–0.90) & 0.90 (0.86–0.93) & 0.90 (0.86–0.93) & 0.90 (0.87–0.94) \\
     & Non-experts & 0.81 (0.73–0.88) & 0.88 (0.83–0.92) & 0.88 (0.83–0.92) & 0.88 (0.84–0.92) \\
    Broad & Gemini 2.5 Pro & 0.78 (0.66–0.89) & 0.90 (0.83–0.95) & 0.90 (0.84–0.96) & 0.90 (0.84–0.96) \\
    ($n=23$) & GPT-5 & 0.64 (0.53–0.76) & 0.78 (0.71–0.84) & 0.77 (0.71–0.84) & 0.79 (0.72–0.85) \\
     & PathChat+ & 0.86 (0.76–0.95) & 0.90 (0.84–0.96) & 0.91 (0.86–0.96) & 0.90 (0.83–0.95) \\
     & Experts & 0.97 (0.92–1.00) & 0.98 (0.95–1.00) & 0.98 (0.95–1.00) & 0.98 (0.94–1.00) \\
     & Non-experts & 0.81 (0.67–0.92) & 0.88 (0.79–0.94) & 0.88 (0.80–0.95) & 0.88 (0.80–0.95) \\
    In-depth & Gemini 2.5 Pro & 0.65 (0.54–0.75) & 0.79 (0.73–0.85) & 0.78 (0.72–0.85) & 0.80 (0.74–0.86) \\
    ($n=20$) & GPT-5 & 0.63 (0.54–0.72) & 0.75 (0.70–0.81) & 0.75 (0.70–0.81) & 0.76 (0.72–0.82) \\
     & PathChat+ & 0.82 (0.70–0.92) & 0.89 (0.82–0.95) & 0.89 (0.82–0.95) & 0.89 (0.82–0.95) \\
     & Experts & 0.91 (0.79–0.98) & 0.92 (0.85–0.98) & 0.92 (0.86–0.98) & 0.92 (0.85–0.98) \\
     & Non-experts & 0.70 (0.55–0.84) & 0.78 (0.68–0.86) & 0.78 (0.68–0.86) & 0.79 (0.70–0.86) \\
    \hline
    \end{tabular}
    \caption{Semantic NLP metric scores for a free-response diagnosis task ($Q_{\text{diagnosis}}$) for DALPHIN$_{reader}$, stratified by diagnostic difficulty (basic [A], broad [B], in-depth knowledge [C], per Dutch pathology education guidelines). 95\% confidence intervals from bootstrapping are included in parentheses.}
    \label{tab:q_diagnosis_difficulty_reader_sem}
\end{table}

\begin{table}[ht]
    \centering
    \begin{tabular}{llllll}
    \hline
    Difficulty & Model/Pathology & BLEU-1 & ROUGE-L & METEOR & CIDEr \\
     & expertise level &  &  &  &  \\
    \hline
    Basic & Gemini 2.5 Pro & 0.50 (0.40–0.60) & 0.53 (0.43–0.63) & 0.45 (0.35–0.55) & 2.84 (2.13–3.65) \\
    ($n=72$) & GPT-5 & 0.41 (0.33–0.51) & 0.46 (0.36–0.56) & 0.36 (0.27–0.45) & 2.36 (1.72–3.04) \\
     & PathChat+ & 0.72 (0.63–0.81) & 0.75 (0.66–0.83) & 0.67 (0.58–0.77) & 4.32 (3.62–5.05) \\
     & Experts & 0.71 (0.63–0.80) & 0.74 (0.66–0.82) & 0.66 (0.57–0.75) & 3.68 (3.00–4.37) \\
     & Non-experts & 0.64 (0.53–0.75) & 0.66 (0.55–0.78) & 0.59 (0.47–0.72) & 3.67 (2.84–4.57) \\
    Broad & Gemini 2.5 Pro & 0.67 (0.48–0.87) & 0.68 (0.48–0.87) & 0.66 (0.46–0.84) & 2.84 (1.75–4.02) \\
    ($n=23$) & GPT-5 & 0.36 (0.19–0.54) & 0.38 (0.21–0.56) & 0.34 (0.18–0.52) & 1.74 (0.81–2.80) \\
     & PathChat+ & 0.68 (0.51–0.84) & 0.72 (0.54–0.87) & 0.65 (0.46–0.83) & 3.31 (2.19–4.56) \\
     & Experts & 0.96 (0.87–1.00) & 0.96 (0.87–1.00) & 0.96 (0.87–1.00) & 4.46 (3.48–5.54) \\
     & Non-experts & 0.62 (0.40–0.82) & 0.66 (0.43–0.86) & 0.58 (0.36–0.79) & 3.02 (1.78–4.47) \\
    In-depth & Gemini 2.5 Pro & 0.27 (0.10–0.47) & 0.27 (0.10–0.47) & 0.26 (0.10–0.46) & 1.30 (0.38–2.43) \\
     ($n=20$) & GPT-5 & 0.23 (0.08–0.40) & 0.24 (0.08–0.42) & 0.19 (0.06–0.36) & 1.06 (0.27–2.02) \\
     & PathChat+ & 0.62 (0.42–0.80) & 0.63 (0.43–0.82) & 0.59 (0.39–0.79) & 2.87 (1.74–3.97) \\
     & Experts & 0.80 (0.62–0.95) & 0.80 (0.63–0.95) & 0.78 (0.60–0.94) & 4.05 (2.92–5.19) \\
     & Non-experts & 0.32 (0.12–0.53) & 0.35 (0.14–0.56) & 0.31 (0.11–0.54) & 1.37 (0.39–2.67) \\
    \hline
    \end{tabular}
    \caption{Traditional NLP metric scores for a free-response diagnosis task ($Q_{\text{diagnosis}}$) for DALPHIN$_{reader}$, stratified by diagnostic difficulty (basic [A], broad [B], in-depth knowledge [C], per Dutch pathology education guidelines). 95\% confidence intervals from bootstrapping are included in parentheses.}
    \label{tab:q_diagnosis_difficulty_reader_trad}
\end{table}

\begin{table}[ht]
    \centering
    \begin{tabular}{llllll}
    \hline
    Incidence & Model & BioBERT sim. & BERTF1 & BERTPrecision & BERTRecall \\
    \hline
    Common cancer & Gemini 2.5 Pro & 0.81 (0.76–0.86) & 0.90 (0.87–0.93) & 0.90 (0.86–0.93) & 0.91 (0.88–0.93) \\
     ($n=65$) & GPT-5 & 0.68 (0.61–0.75) & 0.82 (0.79–0.85) & 0.82 (0.78–0.85) & 0.82 (0.79–0.86) \\
     & PathChat+ & 0.93 (0.89–0.96) & 0.94 (0.91–0.96) & 0.94 (0.93–0.96) & 0.93 (0.91–0.95) \\
    Rare cancer & Gemini 2.5 Pro & 0.74 (0.69–0.79) & 0.85 (0.82–0.88) & 0.85 (0.82–0.88) & 0.85 (0.83–0.88) \\
     ($n=94$) & GPT-5 & 0.69 (0.63–0.74) & 0.81 (0.78–0.84) & 0.81 (0.78–0.84) & 0.81 (0.78–0.84) \\
     & PathChat+ & 0.83 (0.79–0.87) & 0.89 (0.87–0.92) & 0.90 (0.87–0.92) & 0.89 (0.87–0.92) \\
    \hline
    \end{tabular}
    \caption{Semantic NLP metric scores for a free-response diagnosis task ($Q_{\text{diagnosis}}$) for DALPHIN$_{full}$, stratified by cancer incidence (rare cancers <6 per 100,000 per year, per RARECAREnet). 95\% confidence intervals from bootstrapping are included in parentheses.}
    \label{tab:q_diagnosis_cancer_incidence_full_sem}
\end{table}

\begin{table}[ht]
    \centering
    \begin{tabular}{llllll}
    \hline
    Incidence & Model & BLEU-1 & ROUGE-L & METEOR & CIDEr \\
    \hline
    Common cancer & Gemini 2.5 Pro & 0.63 (0.54–0.73) & 0.66 (0.57–0.76) & 0.55 (0.46–0.65) & 3.79 (2.95–4.69) \\
     ($n=65$) & GPT-5 & 0.43 (0.34–0.53) & 0.47 (0.37–0.57) & 0.36 (0.28–0.45) & 2.78 (2.00–3.65) \\
     & PathChat+ & 0.76 (0.70–0.83) & 0.80 (0.73–0.86) & 0.68 (0.59–0.76) & 4.77 (4.03–5.58) \\
    Rare cancer & Gemini 2.5 Pro & 0.54 (0.46–0.63) & 0.56 (0.47–0.64) & 0.48 (0.38–0.56) & 2.90 (2.23–3.60) \\
     ($n=94$) & GPT-5 & 0.42 (0.33–0.51) & 0.45 (0.36–0.53) & 0.38 (0.29–0.47) & 2.46 (1.84–3.15) \\
     & PathChat+ & 0.65 (0.57–0.73) & 0.68 (0.61–0.76) & 0.57 (0.49–0.66) & 3.68 (3.02–4.43) \\
    \hline
    \end{tabular}
    \caption{Traditional NLP metric scores for a free-response diagnosis task ($Q_{\text{diagnosis}}$) for DALPHIN$_{full}$, stratified by cancer incidence (rare cancers <6 per 100,000 per year, per RARECAREnet). 95\% confidence intervals from bootstrapping are included in parentheses.}
    \label{tab:q_diagnosis_cancer_incidence_full_trad}
\end{table}

\begin{table}[ht]
    \centering
    \begin{tabular}{llllll}
    \hline
    Incidence & Model/Pathology & BioBERT sim. & BERTF1 & BERTPrecision & BERTRecall \\
     & expertise level &  &  &  &  \\
    \hline
    Common cancer & Gemini 2.5 Pro & 0.70 (0.60–0.80) & 0.85 (0.80–0.90) & 0.85 (0.79–0.91) & 0.86 (0.82–0.91) \\
     ($n=19$) & GPT-5 & 0.67 (0.52–0.80) & 0.83 (0.76–0.89) & 0.83 (0.76–0.89) & 0.83 (0.77–0.90) \\
     & PathChat+ & 0.96 (0.91–0.99) & 0.96 (0.93–0.98) & 0.97 (0.94–0.99) & 0.96 (0.92–0.98) \\
     & Experts & 0.88 (0.79–0.96) & 0.95 (0.91–0.98) & 0.95 (0.91–0.98) & 0.95 (0.91–0.98) \\
     & Non-experts & 0.86 (0.70–0.97) & 0.92 (0.83–0.98) & 0.92 (0.83–0.98) & 0.92 (0.84–0.98) \\
    Rare cancer & Gemini 2.5 Pro & 0.80 (0.71–0.88) & 0.89 (0.85–0.94) & 0.89 (0.85–0.93) & 0.90 (0.86–0.94) \\
     ($n=33$) & GPT-5 & 0.68 (0.59–0.77) & 0.81 (0.76–0.86) & 0.81 (0.77–0.86) & 0.82 (0.77–0.87) \\
     & PathChat+ & 0.88 (0.81–0.94) & 0.92 (0.88–0.95) & 0.92 (0.88–0.95) & 0.92 (0.88–0.95) \\
     & Experts & 0.96 (0.92–0.99) & 0.97 (0.94–0.99) & 0.97 (0.94–0.99) & 0.97 (0.95–0.99) \\
     & Non-experts & 0.80 (0.70–0.89) & 0.86 (0.79–0.92) & 0.86 (0.79–0.92) & 0.86 (0.80–0.92) \\
    \hline
    \end{tabular}
    \caption{Semantic NLP metric scores for a free-response diagnosis task ($Q_{\text{diagnosis}}$) for DALPHIN$_{reader}$, stratified by cancer incidence (rare cancers <6 per 100,000 per year, per RARECAREnet). 95\% confidence intervals from bootstrapping are included in parentheses.}
    \label{tab:q_diagnosis_cancer_incidence_reader_sem}
\end{table}

\begin{table}[ht]
    \centering
    \begin{tabular}{llllll}
    \hline
    Incidence & Model/Pathology & BLEU-1 & ROUGE-L & METEOR & CIDEr \\
     & expertise level &  &  &  &  \\
    \hline
    Common cancer & Gemini 2.5 Pro & 0.46 (0.29–0.62) & 0.51 (0.33–0.67) & 0.40 (0.24–0.56) & 2.49 (1.03–4.02) \\
     ($n=19$) & GPT-5 & 0.40 (0.21–0.58) & 0.44 (0.25–0.64) & 0.36 (0.20–0.53) & 2.59 (1.20–4.16) \\
     & PathChat+ & 0.85 (0.74–0.95) & 0.88 (0.78–0.96) & 0.81 (0.66–0.93) & 5.52 (4.19–6.80) \\
     & Experts & 0.79 (0.63–0.93) & 0.81 (0.64–0.95) & 0.75 (0.59–0.90) & 5.16 (3.43–6.92) \\
     & Non-experts & 0.72 (0.49–0.92) & 0.74 (0.50–0.93) & 0.67 (0.43–0.89) & 4.53 (2.64–6.39) \\
    Rare cancer & Gemini 2.5 Pro & 0.65 (0.52–0.79) & 0.67 (0.53–0.80) & 0.60 (0.45–0.75) & 3.46 (2.38–4.54) \\
     ($n=33$) & GPT-5 & 0.42 (0.27–0.56) & 0.44 (0.29–0.59) & 0.37 (0.24–0.51) & 2.39 (1.39–3.45) \\
     & PathChat+ & 0.71 (0.58–0.83) & 0.76 (0.64–0.87) & 0.64 (0.50–0.78) & 3.83 (2.80–4.84) \\
     & Experts & 0.91 (0.84–0.98) & 0.92 (0.85–0.98) & 0.87 (0.76–0.96) & 4.63 (3.67–5.56) \\
     & Non-experts & 0.57 (0.39–0.75) & 0.60 (0.44–0.78) & 0.51 (0.33–0.71) & 3.23 (1.95–4.66) \\
    \hline
    \end{tabular}
    \caption{Traditional NLP metric scores for a free-response diagnosis task ($Q_{\text{diagnosis}}$) for DALPHIN$_{reader}$, stratified by cancer incidence (rare cancers <6 per 100,000 per year, per RARECAREnet). 95\% confidence intervals from bootstrapping are included in parentheses.}
    \label{tab:q_diagnosis_cancer_incidence_reader_trad}
\end{table}

\begin{table}[ht]
    \centering
    \begin{tabular}{llllll}
    \hline
    Incidence & Model & BioBERT sim. & BERTF1 & BERTPrecision & BERTRecall \\
    \hline
    Neoplastic & Gemini 2.5 Pro & 0.74 (0.71–0.78) & 0.85 (0.83–0.87) & 0.85 (0.83–0.87) & 0.85 (0.84–0.87) \\
     ($n=232$) & GPT-5 & 0.67 (0.63–0.70) & 0.80 (0.78–0.82) & 0.80 (0.78–0.82) & 0.80 (0.79–0.82) \\
     & PathChat+ & 0.86 (0.83–0.89) & 0.91 (0.89–0.92) & 0.91 (0.90–0.93) & 0.90 (0.89–0.92) \\
    Non-neoplastic & Gemini 2.5 Pro & 0.60 (0.52–0.66) & 0.77 (0.73–0.81) & 0.77 (0.73–0.81) & 0.78 (0.74–0.82) \\
     ($n=68$) & GPT-5 & 0.65 (0.58–0.72) & 0.77 (0.73–0.81) & 0.77 (0.73–0.80) & 0.78 (0.74–0.82) \\
     & PathChat+ & 0.80 (0.74–0.86) & 0.85 (0.81–0.89) & 0.85 (0.81–0.89) & 0.85 (0.81–0.89) \\
    \hline
    \end{tabular}
    \caption{Semantic NLP metric scores for a free-response diagnosis task ($Q_{\text{diagnosis}}$) for DALPHIN$_{full}$, stratified by neoplastic status. 95\% confidence intervals from bootstrapping are included in parentheses.}
    \label{tab:q_diagnosis_neoplastic_status_full_sem}
\end{table}

\begin{table}[ht]
    \centering
    \begin{tabular}{llllll}
    \hline
    Incidence & Model & BLEU-1 & ROUGE-L & METEOR & CIDEr \\
    \hline
    Neoplastic & Gemini 2.5 Pro & 0.54 (0.48–0.60) & 0.56 (0.50–0.62) & 0.48 (0.42–0.54) & 2.96 (2.53–3.40) \\
     ($n=232$) & GPT-5 & 0.39 (0.33–0.44) & 0.42 (0.36–0.47) & 0.34 (0.29–0.40) & 2.32 (1.89–2.74) \\
     & PathChat+ & 0.69 (0.64–0.74) & 0.73 (0.68–0.77) & 0.63 (0.57–0.68) & 4.07 (3.63–4.47) \\
    Non-neoplastic & Gemini 2.5 Pro & 0.31 (0.20–0.42) & 0.32 (0.21–0.42) & 0.28 (0.18–0.38) & 1.67 (1.01–2.36) \\
     ($n=68$) & GPT-5 & 0.35 (0.26–0.44) & 0.38 (0.28–0.47) & 0.28 (0.19–0.37) & 1.61 (1.06–2.22) \\
     & PathChat+ & 0.54 (0.43–0.65) & 0.55 (0.44–0.67) & 0.51 (0.40–0.63) & 2.80 (2.13–3.54) \\
    \hline
    \end{tabular}
    \caption{Traditional NLP metric scores for a free-response diagnosis task ($Q_{\text{diagnosis}}$) for DALPHIN$_{full}$, stratified by neoplastic status. 95\% confidence intervals from bootstrapping are included in parentheses.}
    \label{tab:q_diagnosis_neoplastic_status_full_trad}
\end{table}

\begin{table}[ht]
    \centering
    \begin{tabular}{llllll}
    \hline
    Incidence & Model/Pathology & BioBERT sim. & BERTF1 & BERTPrecision & BERTRecall \\
     & expertise level &  &  &  &  \\
    \hline
    Neoplastic & Gemini 2.5 Pro & 0.76 (0.70–0.81) & 0.87 (0.84–0.90) & 0.87 (0.84–0.90) & 0.87 (0.84–0.90) \\
     ($n=79$) & GPT-5 & 0.69 (0.62–0.75) & 0.81 (0.78–0.84) & 0.81 (0.78–0.84) & 0.82 (0.78–0.85) \\
     & PathChat+ & 0.90 (0.85–0.94) & 0.93 (0.91–0.95) & 0.94 (0.91–0.96) & 0.93 (0.90–0.95) \\
     & Experts & 0.93 (0.90–0.96) & 0.96 (0.94–0.98) & 0.96 (0.94–0.98) & 0.96 (0.94–0.98) \\
     & Non-experts & 0.82 (0.75–0.88) & 0.88 (0.84–0.92) & 0.88 (0.84–0.92) & 0.88 (0.84–0.92) \\
    Non-neoplastic & Gemini 2.5 Pro & 0.62 (0.53–0.72) & 0.79 (0.75–0.84) & 0.79 (0.74–0.84) & 0.80 (0.76–0.85) \\
     ($n=36$) & GPT-5 & 0.61 (0.53–0.69) & 0.75 (0.70–0.79) & 0.74 (0.69–0.78) & 0.76 (0.72–0.81) \\
     & PathChat+ & 0.79 (0.70–0.86) & 0.86 (0.81–0.91) & 0.86 (0.81–0.90) & 0.87 (0.81–0.91) \\
     & Experts & 0.76 (0.66–0.86) & 0.83 (0.77–0.89) & 0.83 (0.77–0.90) & 0.84 (0.77–0.90) \\
     & Non-experts & 0.73 (0.62–0.85) & 0.82 (0.75–0.89) & 0.82 (0.75–0.89) & 0.83 (0.77–0.89) \\
    \hline
    \end{tabular}
    \caption{Semantic NLP metric scores for a free-response diagnosis task ($Q_{\text{diagnosis}}$) for DALPHIN$_{reader}$, stratified by neoplastic status. 95\% confidence intervals from bootstrapping are included in parentheses.}
    \label{tab:q_diagnosis_neoplastic_status_reader_sem}
\end{table}

\begin{table}[ht]
    \centering
    \begin{tabular}{llllll}
    \hline
    Incidence & Model/Pathology & BLEU-1 & ROUGE-L & METEOR & CIDEr \\
     & expertise level &  &  &  &  \\
    \hline
    Neoplastic & Gemini 2.5 Pro & 0.57 (0.48–0.66) & 0.59 (0.50–0.68) & 0.53 (0.43–0.63) & 2.97 (2.28–3.69) \\
     ($n=79$) & GPT-5 & 0.41 (0.31–0.50) & 0.45 (0.34–0.54) & 0.38 (0.28–0.47) & 2.37 (1.72–3.04) \\
     & PathChat+ & 0.77 (0.69–0.84) & 0.80 (0.73–0.87) & 0.72 (0.64–0.81) & 4.40 (3.79–5.07) \\
     & Experts & 0.87 (0.81–0.92) & 0.89 (0.83–0.94) & 0.83 (0.76–0.90) & 4.52 (3.86–5.20) \\
     & Non-experts & 0.63 (0.52–0.74) & 0.66 (0.54–0.76) & 0.59 (0.47–0.71) & 3.60 (2.70–4.50) \\
    Non-neoplastic & Gemini 2.5 Pro & 0.33 (0.19–0.49) & 0.34 (0.20–0.49) & 0.30 (0.17–0.45) & 1.70 (0.91–2.58) \\
     ($n=36$) & GPT-5 & 0.28 (0.17–0.40) & 0.31 (0.20–0.44) & 0.21 (0.11–0.32) & 1.21 (0.62–1.97) \\
     & PathChat+ & 0.54 (0.39–0.68) & 0.55 (0.41–0.69) & 0.51 (0.36–0.65) & 2.70 (1.83–3.50) \\
     & Experts & 0.58 (0.42–0.73) & 0.60 (0.44–0.74) & 0.55 (0.40–0.70) & 2.55 (1.77–3.41) \\
     & Non-experts & 0.47 (0.31–0.66) & 0.49 (0.33–0.68) & 0.43 (0.27–0.63) & 2.14 (1.26–3.21) \\
    \hline
    \end{tabular}
    \caption{Traditional NLP metric scores for a free-response diagnosis task ($Q_{\text{diagnosis}}$) for DALPHIN$_{reader}$, stratified by neoplastic status. 95\% confidence intervals from bootstrapping are included in parentheses.}
    \label{tab:q_diagnosis_neoplastic_status_reader_trad}
\end{table}

\begin{table}[ht]
    \centering
    \begin{tabular}{llllll}
    \hline
    Subspecialty & Model & BioBERT sim. & BERTF1 & BERTPrecision & BERTRecall \\
    \hline
    Genitourinary & Gemini 2.5 Pro & 0.83 (0.78–0.88) & 0.88 (0.85–0.92) & 0.89 (0.86–0.92) & 0.88 (0.85–0.92) \\
    ($n=59$) & GPT-5 & 0.78 (0.72–0.83) & 0.83 (0.79–0.87) & 0.84 (0.80–0.87) & 0.83 (0.79–0.87) \\
     & PathChat+ & 0.89 (0.84–0.93) & 0.90 (0.87–0.93) & 0.91 (0.88–0.94) & 0.89 (0.85–0.92) \\
    Gastrointestinal & Gemini 2.5 Pro & 0.57 (0.49–0.63) & 0.77 (0.73–0.81) & 0.77 (0.72–0.81) & 0.79 (0.74–0.83) \\
    ($n=52$) & GPT-5 & 0.65 (0.58–0.73) & 0.79 (0.75–0.83) & 0.78 (0.74–0.82) & 0.80 (0.76–0.85) \\
     & PathChat+ & 0.86 (0.80–0.91) & 0.90 (0.87–0.93) & 0.90 (0.87–0.94) & 0.90 (0.86–0.93) \\
    Soft tissue & Gemini 2.5 Pro & 0.60 (0.53–0.67) & 0.75 (0.71–0.79) & 0.76 (0.72–0.80) & 0.75 (0.70–0.79) \\
\hline
    \end{tabular}
    \caption{Semantic NLP metric scores for a free-response diagnosis task ($Q_{\text{diagnosis}}$) for DALPHIN$_{full}$, stratified by subspecialty. Only subspecialties with $\geq$10 cases are shown. 95\% confidence intervals from bootstrapping are included in parentheses.}
    \label{tab:q_diagnosis_subspecialty_sem}
\end{table}

\begin{table}[ht]
    \centering
    \begin{tabular}{llllll}
    \hline
    Subspecialty & Model & BLEU-1 & ROUGE-L & METEOR & CIDEr \\
    \hline
    Genitourinary & Gemini 2.5 Pro & 0.64 (0.54–0.74) & 0.66 (0.57–0.76) & 0.58 (0.47–0.69) & 4.21 (3.25–5.25) \\
    ($n=59$)  & GPT-5 & 0.51 (0.41–0.62) & 0.56 (0.46–0.66) & 0.44 (0.34–0.56) & 3.34 (2.39–4.34) \\
     & PathChat+ & 0.67 (0.59–0.76) & 0.70 (0.61–0.78) & 0.55 (0.45–0.65) & 4.40 (3.47–5.48) \\
    Gastrointestinal & Gemini 2.5 Pro & 0.34 (0.23–0.44) & 0.35 (0.24–0.46) & 0.30 (0.19–0.40) & 1.45 (0.83–2.11) \\
    ($n=52$)  & GPT-5 & 0.36 (0.26–0.47) & 0.40 (0.29–0.52) & 0.30 (0.20–0.41) & 1.92 (1.16–2.73) \\
     & PathChat+ & 0.66 (0.56–0.76) & 0.71 (0.61–0.80) & 0.60 (0.50–0.71) & 3.67 (2.91–4.49) \\
    Soft tissue & Gemini 2.5 Pro & 0.26 (0.15–0.38) & 0.26 (0.15–0.38) & 0.25 (0.13–0.37) & 1.17 (0.58–1.84) \\
\hline
    \end{tabular}
    \caption{Traditional NLP metric scores for a free-response diagnosis task ($Q_{\text{diagnosis}}$) for DALPHIN$_{full}$, stratified by subspecialty. Only subspecialties with $\geq$10 cases are shown. 95\% confidence intervals from bootstrapping are included in parentheses.}
    \label{tab:q_diagnosis_subspecialty_trad}
\end{table}

\begin{table}[ht]
    \centering
    \begin{tabular}{llll}
    \hline
    Model/Pathology & DALPHIN$_{full}$ & DALPHIN$_{reader}$ & DALPHIN$_{semi}$ \\
    expertise level & ($n=300$) & ($n=115$) & ($n=60$) \\
    \hline
    Gemini 2.5 Pro & 0.757 (0.707–0.803) & 0.704 (0.617–0.783) & 0.750 (0.633–0.867) \\
    GPT-5 & 0.743 (0.693–0.790) & 0.783 (0.713–0.861) & 0.733 (0.617–0.833) \\
    PathChat+ & 0.777 (0.730–0.820) & 0.765 (0.678–0.843) & 0.767 (0.650–0.867) \\
    Experts & NA & 0.843 (0.783–0.913) & 0.817 (0.717–0.917) \\
    Non-experts & NA & 0.755 (0.652–0.852) & 0.795 (0.667–0.917) \\
    Semi-experts & NA & NA & 0.799 (0.667–0.917) \\
    \hline
    \end{tabular}
    \caption{Accuracy on case-specific multiple-choice questions ($Q_{\text{mc}}$). 95\% confidence intervals from bootstrapping are included in parentheses.}
    \label{tab:q_mc}
\end{table}

\begin{table}[ht]
    \centering
    \begin{tabular}{lll}
    \hline
    Category & Model & DALPHIN$_{full}$ \\
    \hline
    Morphology ($n=94$) & Gemini 2.5 Pro & 0.777 (0.691–0.862) \\
     & GPT-5 & 0.777 (0.691–0.862) \\
     & PathChat+ & 0.702 (0.606–0.787) \\
    \hline
    Mix ($n=75$) & Gemini 2.5 Pro & 0.733 (0.640–0.827) \\
     & GPT-5 & 0.680 (0.573–0.787) \\
     & PathChat+ & 0.813 (0.733–0.893) \\
    \hline
    Ancillary testing ($n=60$) & Gemini 2.5 Pro & 0.833 (0.733–0.917) \\
     & GPT-5 & 0.750 (0.633–0.850) \\
     & PathChat+ & 0.833 (0.733–0.917) \\
    \hline
    Clinical ($n=46$) & Gemini 2.5 Pro & 0.739 (0.609–0.848) \\
     & GPT-5 & 0.783 (0.674–0.891) \\
     & PathChat+ & 0.783 (0.652–0.891) \\
    \hline
    Diagnosis ($n=25$) & Gemini 2.5 Pro & 0.600 (0.400–0.800) \\
     & GPT-5 & 0.720 (0.520–0.880) \\
     & PathChat+ & 0.800 (0.640–0.920) \\
    \hline
    \end{tabular}
    \caption{Accuracy on case-specific multiple-choice questions ($Q_{\text{mc}}$), stratified by question category. 95\% confidence intervals from bootstrapping are included in parentheses.}
    \label{tab:q_mc_category}
\end{table}

\begin{table}[ht]
    \centering
    \begin{tabular}{llllll}
    \hline
    Subset & Model/Pathology & BioBERT sim. & BERTF1 & BERTPrecision & BERTRecall \\
     & expertise level &  &  &  &  \\
    \hline
    DALPHIN$_{full}$ & Gemini 2.5 Pro & 0.64 (0.61–0.66) & 0.68 (0.66–0.70) & 0.70 (0.68–0.71) & 0.68 (0.66–0.70) \\
    ($n=285$) & GPT-5 & 0.62 (0.59–0.65) & 0.65 (0.64–0.67) & 0.63 (0.61–0.64) & 0.68 (0.67–0.70) \\
     & PathChat+ & 0.62 (0.59–0.64) & 0.64 (0.63–0.66) & 0.64 (0.62–0.65) & 0.66 (0.64–0.67) \\
    DALPHIN$_{reader}$ & Gemini 2.5 Pro & 0.63 (0.58–0.67) & 0.67 (0.64–0.70) & 0.69 (0.66–0.72) & 0.66 (0.63–0.70) \\
    ($n=105$) & GPT-5 & 0.61 (0.57–0.65) & 0.65 (0.62–0.67) & 0.63 (0.61–0.66) & 0.68 (0.65–0.70) \\
     & PathChat+ & 0.64 (0.60–0.67) & 0.64 (0.62–0.67) & 0.64 (0.61–0.67) & 0.66 (0.63–0.69) \\
     & Experts & 0.64 (0.59–0.68) & 0.68 (0.65–0.71) & 0.72 (0.69–0.75) & 0.66 (0.63–0.70) \\
     & Non-experts & 0.62 (0.57–0.68) & 0.67 (0.63–0.71) & 0.69 (0.66–0.73) & 0.65 (0.61–0.69) \\
    DALPHIN$_{semi}$ & Gemini 2.5 Pro & 0.63 (0.58–0.69) & 0.67 (0.63–0.71) & 0.69 (0.65–0.73) & 0.66 (0.62–0.70) \\
    ($n=54$) & GPT-5 & 0.62 (0.56–0.67) & 0.66 (0.62–0.69) & 0.64 (0.60–0.68) & 0.68 (0.65–0.72) \\
     & PathChat+ & 0.66 (0.60–0.71) & 0.66 (0.62–0.70) & 0.66 (0.61–0.70) & 0.67 (0.63–0.71) \\
     & Experts & 0.63 (0.57–0.68) & 0.68 (0.65–0.72) & 0.72 (0.69–0.75) & 0.66 (0.62–0.70) \\
     & Non-experts & 0.63 (0.55–0.70) & 0.67 (0.62–0.72) & 0.70 (0.65–0.75) & 0.65 (0.60–0.70) \\
     & Semi-experts & 0.63 (0.56–0.69) & 0.67 (0.63–0.70) & 0.70 (0.66–0.74) & 0.64 (0.60–0.68) \\
    \hline
    \end{tabular}
    \caption{Semantic NLP metric scores for case-specific free-response questions ($Q_{\text{open}}$). 95\% confidence intervals from bootstrapping are included in parentheses.}
    \label{tab:q_open_sem}
\end{table}

\begin{table}[ht]
    \centering
    \begin{tabular}{llllll}
    \hline
    Subset & Model/Pathology & BLEU-1 & ROUGE-L & METEOR & CIDEr \\
     & expertise level &  &  &  &  \\
    \hline
    DALPHIN$_{full}$ & Gemini 2.5 Pro & 0.24 (0.20–0.27) & 0.27 (0.23–0.30) & 0.25 (0.21–0.28) & 0.82 (0.67–0.98) \\
    ($n=285$) & GPT-5 & 0.18 (0.15–0.20) & 0.21 (0.18–0.23) & 0.19 (0.17–0.21) & 0.57 (0.46–0.70) \\
     & PathChat+ & 0.19 (0.16–0.22) & 0.23 (0.19–0.26) & 0.21 (0.18–0.23) & 0.61 (0.45–0.75) \\
    DALPHIN$_{reader}$ & Gemini 2.5 Pro & 0.21 (0.16–0.27) & 0.24 (0.19–0.30) & 0.23 (0.18–0.28) & 0.64 (0.43–0.89) \\
    ($n=105$) & GPT-5 & 0.19 (0.15–0.24) & 0.23 (0.18–0.28) & 0.19 (0.16–0.23) & 0.60 (0.41–0.79) \\
     & PathChat+ & 0.19 (0.14–0.24) & 0.23 (0.18–0.29) & 0.21 (0.17–0.26) & 0.60 (0.36–0.88) \\
     & Experts & 0.23 (0.17–0.29) & 0.27 (0.21–0.33) & 0.24 (0.19–0.31) & 0.77 (0.55–1.03) \\
     & Non-experts & 0.20 (0.14–0.26) & 0.24 (0.18–0.31) & 0.21 (0.15–0.27) & 0.73 (0.47–1.04) \\
    DALPHIN$_{semi}$ & Gemini 2.5 Pro & 0.20 (0.14–0.27) & 0.23 (0.17–0.31) & 0.20 (0.14–0.28) & 0.53 (0.32–0.78) \\
    ($n=54$) & GPT-5 & 0.20 (0.15–0.27) & 0.23 (0.17–0.30) & 0.19 (0.14–0.26) & 0.53 (0.33–0.80) \\
     & PathChat+ & 0.20 (0.14–0.29) & 0.24 (0.17–0.32) & 0.23 (0.17–0.31) & 0.61 (0.32–0.98) \\
     & Experts & 0.19 (0.13–0.27) & 0.24 (0.17–0.32) & 0.20 (0.14–0.27) & 0.63 (0.38–0.92) \\
     & Non-experts & 0.19 (0.11–0.29) & 0.24 (0.16–0.34) & 0.20 (0.13–0.30) & 0.74 (0.39–1.29) \\
     & Semi-experts & 0.17 (0.10–0.24) & 0.21 (0.14–0.30) & 0.17 (0.11–0.24) & 0.59 (0.31–0.97) \\
    \hline
    \end{tabular}
    \caption{Traditional NLP metric scores for case-specific free-response questions ($Q_{\text{open}}$). 95\% confidence intervals from bootstrapping are included in parentheses.}
    \label{tab:q_open_trad}
\end{table}

\begin{table}[ht]
    \centering
    \begin{tabular}{llllll}
    \hline
    Category & Model & BioBERT sim. & BERTF1 & BERTPrecision & BERTRecall \\
    \hline
    Morphology & Gemini 2.5 Pro & 0.63 (0.59–0.68) & 0.68 (0.65–0.71) & 0.68 (0.66–0.71) & 0.68 (0.66–0.71) \\
    ($n=108$) & GPT-5 & 0.63 (0.59–0.67) & 0.66 (0.64–0.69) & 0.64 (0.62–0.67) & 0.69 (0.67–0.72) \\
     & PathChat+ & 0.65 (0.62–0.68) & 0.65 (0.63–0.67) & 0.64 (0.62–0.66) & 0.67 (0.65–0.69) \\
    Ancillary testing & Gemini 2.5 Pro & 0.59 (0.52–0.66) & 0.68 (0.64–0.72) & 0.71 (0.67–0.75) & 0.66 (0.62–0.70) \\
    ($n=67$) & GPT-5 & 0.59 (0.54–0.64) & 0.62 (0.59–0.65) & 0.60 (0.57–0.63) & 0.65 (0.62–0.68) \\
     & PathChat+ & 0.56 (0.50–0.63) & 0.63 (0.59–0.67) & 0.64 (0.60–0.68) & 0.63 (0.59–0.67) \\
    Clinical & Gemini 2.5 Pro & 0.71 (0.65–0.78) & 0.72 (0.67–0.77) & 0.74 (0.69–0.79) & 0.71 (0.66–0.76) \\
     ($n=55$) & GPT-5 & 0.60 (0.53–0.67) & 0.66 (0.62–0.71) & 0.64 (0.59–0.68) & 0.70 (0.66–0.74) \\
     & PathChat+ & 0.58 (0.50–0.65) & 0.63 (0.59–0.68) & 0.63 (0.58–0.68) & 0.65 (0.60–0.70) \\
    Diagnosis & Gemini 2.5 Pro & 0.64 (0.58–0.69) & 0.66 (0.62–0.70) & 0.66 (0.62–0.70) & 0.66 (0.62–0.70) \\
     ($n=55$) & GPT-5 & 0.65 (0.59–0.70) & 0.65 (0.63–0.68) & 0.62 (0.60–0.65) & 0.69 (0.66–0.72) \\
     & PathChat+ & 0.65 (0.61–0.70) & 0.65 (0.62–0.68) & 0.64 (0.61–0.68) & 0.67 (0.64–0.69) \\
    \hline
    \end{tabular}
    \caption{Semantic NLP metric scores for case-specific free-response questions ($Q_{\text{open}}$) for DALPHIN$_{full}$, stratified by question category. 95\% confidence intervals from bootstrapping are included in parentheses.}
    \label{tab:q_open_category_sem}
\end{table}

\begin{table}[ht]
    \centering
    \begin{tabular}{llllll}
    \hline
    Category & Model & BLEU-1 & ROUGE-L & METEOR & CIDEr \\
    \hline
    Morphology & Gemini 2.5 Pro & 0.22 (0.16–0.27) & 0.25 (0.20–0.31) & 0.24 (0.18–0.30) & 0.63 (0.41–0.88) \\
     ($n=108$) & GPT-5 & 0.21 (0.17–0.26) & 0.24 (0.19–0.28) & 0.20 (0.16–0.24) & 0.60 (0.40–0.80) \\
     & PathChat+ & 0.20 (0.16–0.25) & 0.25 (0.20–0.30) & 0.22 (0.18–0.27) & 0.54 (0.34–0.78) \\
    Ancillary testing & Gemini 2.5 Pro & 0.24 (0.16–0.31) & 0.27 (0.20–0.35) & 0.24 (0.18–0.31) & 0.79 (0.52–1.07) \\
     ($n=67$) & GPT-5 & 0.15 (0.11–0.20) & 0.19 (0.14–0.25) & 0.18 (0.14–0.22) & 0.55 (0.33–0.82) \\
     & PathChat+ & 0.17 (0.11–0.24) & 0.20 (0.14–0.27) & 0.19 (0.14–0.26) & 0.57 (0.33–0.86) \\
    Clinical & Gemini 2.5 Pro & 0.34 (0.24–0.44) & 0.37 (0.28–0.48) & 0.34 (0.24–0.43) & 1.39 (0.95–1.86) \\
     ($n=55$) & GPT-5 & 0.18 (0.12–0.26) & 0.21 (0.14–0.30) & 0.21 (0.15–0.27) & 0.71 (0.41–1.10) \\
     & PathChat+ & 0.21 (0.13–0.29) & 0.25 (0.17–0.33) & 0.23 (0.16–0.30) & 0.79 (0.43–1.20) \\
    Diagnosis & Gemini 2.5 Pro & 0.19 (0.13–0.26) & 0.20 (0.14–0.27) & 0.19 (0.13–0.25) & 0.69 (0.36–1.06) \\
     ($n=55$) & GPT-5 & 0.14 (0.10–0.18) & 0.16 (0.12–0.21) & 0.16 (0.13–0.19) & 0.42 (0.23–0.64) \\
     & PathChat+ & 0.17 (0.12–0.23) & 0.20 (0.14–0.26) & 0.18 (0.13–0.23) & 0.59 (0.27–1.07) \\
    \hline
    \end{tabular}
    \caption{Traditional NLP metric scores for case-specific free-response questions ($Q_{\text{open}}$) for DALPHIN$_{full}$, stratified by question category. 95\% confidence intervals from bootstrapping are included in parentheses.}
    \label{tab:q_open_category_trad}
\end{table}

\begin{table}[ht]
    \centering
    \begin{tabular}{llllll}
    \hline
    Subspecialty & Model & BioBERT sim. & BERTF1 & BERTPrecision & BERTRecall \\
    \hline
    Genitourinary & Gemini 2.5 Pro & 0.66 (0.58–0.72) & 0.70 (0.66–0.75) & 0.71 (0.66–0.75) & 0.71 (0.66–0.75) \\
    ($n=55$) & GPT-5 & 0.57 (0.51–0.63) & 0.63 (0.60–0.66) & 0.59 (0.56–0.63) & 0.68 (0.65–0.71) \\
     & PathChat+ & 0.52 (0.46–0.58) & 0.60 (0.57–0.63) & 0.60 (0.57–0.63) & 0.62 (0.58–0.65) \\
    Gastrointestinal & Gemini 2.5 Pro & 0.56 (0.50–0.62) & 0.62 (0.59–0.65) & 0.64 (0.61–0.67) & 0.61 (0.58–0.65) \\
     ($n=51$) & GPT-5 & 0.61 (0.55–0.67) & 0.63 (0.60–0.66) & 0.62 (0.59–0.64) & 0.65 (0.61–0.68) \\
     & PathChat+ & 0.64 (0.58–0.69) & 0.63 (0.60–0.67) & 0.63 (0.60–0.66) & 0.64 (0.61–0.68) \\
    Soft tissue & Gemini 2.5 Pro & 0.56 (0.49–0.63) & 0.64 (0.60–0.69) & 0.65 (0.61–0.69) & 0.64 (0.60–0.68) \\
\hline
    \end{tabular}
    \caption{Semantic NLP metric scores for case-specific free-response questions ($Q_{\text{open}}$) for DALPHIN$_{full}$, stratified by subspecialty. Only subspecialties with $\geq$10 cases are shown. 95\% confidence intervals from bootstrapping are included in parentheses.}
    \label{tab:q_open_subspecialty_sem}
\end{table}

\begin{table}[ht]
    \centering
    \begin{tabular}{llllll}
    \hline
    Subspecialty & Model & BLEU-1 & ROUGE-L & METEOR & CIDEr \\
    \hline
    Genitourinary & Gemini 2.5 Pro & 0.28 (0.19–0.38) & 0.32 (0.23–0.41) & 0.30 (0.22–0.40) & 0.97 (0.60–1.39) \\
     ($n=55$) & GPT-5 & 0.17 (0.12–0.22) & 0.20 (0.14–0.26) & 0.18 (0.14–0.21) & 0.61 (0.35–0.96) \\
     & PathChat+ & 0.16 (0.11–0.23) & 0.20 (0.14–0.27) & 0.17 (0.12–0.22) & 0.55 (0.32–0.81) \\
    Gastrointestinal & Gemini 2.5 Pro & 0.15 (0.10–0.20) & 0.18 (0.13–0.24) & 0.15 (0.11–0.19) & 0.47 (0.25–0.73) \\
     ($n=51$) & GPT-5 & 0.14 (0.11–0.18) & 0.16 (0.12–0.21) & 0.16 (0.12–0.19) & 0.35 (0.19–0.54) \\
     & PathChat+ & 0.18 (0.13–0.25) & 0.21 (0.15–0.27) & 0.19 (0.14–0.25) & 0.53 (0.25–0.84) \\
    Soft tissue & Gemini 2.5 Pro & 0.15 (0.08–0.23) & 0.18 (0.11–0.26) & 0.17 (0.09–0.25) & 0.48 (0.26–0.72) \\
\hline
    \end{tabular}
    \caption{Traditional NLP metric scores for case-specific free-response questions ($Q_{\text{open}}$) for DALPHIN$_{full}$, stratified by subspecialty. Only subspecialties with $\geq$10 cases are shown. 95\% confidence intervals from bootstrapping are included in parentheses.}
    \label{tab:q_open_subspecialty_trad}
\end{table}

\begin{table}[ht]
    \centering
    \begin{tabular}{lllll}
    \hline
    Task & Metric & Model & Contextual & Independent \\
    \hline
    Q$_{tissue}$ & Organ recognition score & Gemini 2.5 Pro & 0.734 (0.683–0.781) & 0.719 (0.668–0.769) \\
    ($n=272$) &  & GPT-5 & 0.436 (0.378–0.495) & 0.449 (0.394–0.504) \\
     &  & PathChat+ & 0.787 (0.737–0.836) & 0.767 (0.716–0.819) \\
    \hline
    Q$_{neoplasm}$ & F1 & Gemini 2.5 Pro & 0.925 (0.901–0.947) & 0.917 (0.891–0.940) \\
    ($n=300$) &  & GPT-5 & 0.844 (0.803–0.879) & 0.814 (0.768–0.854) \\
     &  & PathChat+ & 0.968 (0.951–0.982)** & 0.945 (0.922–0.965) \\
     & MCC & Gemini 2.5 Pro & 0.620 (0.505–0.713) & 0.572 (0.457–0.676) \\
\hline
    \end{tabular}
    \caption{Performance on initial case-orienting tasks for DALPHIN$_{full}$, comparing the contextual (feedback/sequential) and independent answer generation scenarios (see `VLM evaluation' in Methods). 95\% confidence intervals from bootstrapping are included in parentheses. *$P$<0.05; **$P$<0.01.}
    \label{tab:q_all_i_contextual_independent}
\end{table}

\begin{table}[ht]
    \centering
    \begin{tabular}{llll}
    \hline
    Metric & Model & Contextual & Independent \\
    \hline
    BioBERT similarity & Gemini 2.5 Pro & 0.709 (0.678–0.738) & 0.714 (0.682–0.743) \\
     & GPT-5 & 0.663 (0.631–0.694) & 0.639 (0.608–0.671) \\
     & PathChat+ & 0.846 (0.821–0.874) & 0.827 (0.802–0.855) \\
     BERTF1 & Gemini 2.5 Pro & 0.833 (0.816–0.850) & 0.832 (0.814–0.849) \\
     & GPT-5 & 0.792 (0.775–0.809) & 0.778 (0.762–0.795) \\
     & PathChat+ & 0.893 (0.878–0.908) & 0.882 (0.866–0.898) \\
     BERTPrecision & Gemini 2.5 Pro & 0.834 (0.817–0.851) & 0.834 (0.815–0.851) \\
     & GPT-5 & 0.792 (0.776–0.809) & 0.775 (0.758–0.793) \\
     & PathChat+ & 0.897 (0.882–0.913) & 0.889 (0.874–0.905) \\
     BERTRecall & Gemini 2.5 Pro & 0.838 (0.821–0.854) & 0.837 (0.820–0.854) \\
     & GPT-5 & 0.800 (0.783–0.817) & 0.790 (0.774–0.807) \\
     & PathChat+ & 0.893 (0.878–0.908)* & 0.880 (0.865–0.897) \\
     BLEU-1 & Gemini 2.5 Pro & 0.488 (0.437–0.537) & 0.495 (0.445–0.542) \\
     & GPT-5 & 0.377 (0.328–0.422) & 0.346 (0.303–0.393) \\
     & PathChat+ & 0.660 (0.614–0.708) & 0.628 (0.586–0.678) \\
     ROUGE-L & Gemini 2.5 Pro & 0.505 (0.454–0.552) & 0.516 (0.465–0.564) \\
     & GPT-5 & 0.409 (0.358–0.455) & 0.376 (0.331–0.424) \\
     & PathChat+ & 0.686 (0.641–0.735) & 0.661 (0.617–0.710) \\
     METEOR & Gemini 2.5 Pro & 0.435 (0.385–0.486) & 0.442 (0.391–0.491) \\
     & GPT-5 & 0.330 (0.283–0.375) & 0.291 (0.251–0.336) \\
     & PathChat+ & 0.600 (0.554–0.651) & 0.566 (0.523–0.617) \\
     CIDEr & Gemini 2.5 Pro & 2.666 (2.303–3.049) & 2.602 (2.251–2.942) \\
     & GPT-5 & 2.160 (1.806–2.504) & 1.868 (1.558–2.206) \\
     & PathChat+ & 3.782 (3.433–4.166) & 3.555 (3.195–3.937) \\
    \hline
    \end{tabular}
    \caption{Performance on a free-response diagnosis task (Q$_{diagnosis}$) for DALPHIN$_{full}$ ($n=300$), comparing the contextual (feedback/sequential) and independent answer generation scenarios (see `VLM evaluation' in Methods). 95\% confidence intervals from bootstrapping are included in parentheses. *$P$<0.05; **$P$<0.01.}
    \label{tab:q_all_ii_contextual_independent}
\end{table}

\begin{table}[ht]
    \centering
    \begin{tabular}{lllll}
    \hline
    Task & Metric & Model & Contextual & Independent \\
    \hline
    Q$_{mc}$ & Accuracy & Gemini 2.5 Pro & 0.757 (0.707–0.803) & 0.840 (0.797–0.880)** \\
    ($n=300$) &  & GPT-5 & 0.743 (0.693–0.790) & 0.807 (0.760–0.857)* \\
     &  & PathChat+ & 0.777 (0.730–0.820)* & 0.737 (0.683–0.783) \\
    \hline
    Q$_{open}$ & BioBERT similarity & Gemini 2.5 Pro & 0.638 (0.608–0.665) & 0.648 (0.619–0.677) \\
    ($n=285$) &  & GPT-5 & 0.618 (0.591–0.646) & 0.610 (0.585–0.636) \\
     &  & PathChat+ & 0.616 (0.590–0.642)* & 0.592 (0.565–0.621) \\
     & BERTF1 & Gemini 2.5 Pro & 0.682 (0.664–0.701) & 0.687 (0.669–0.703) \\
\hline
    \end{tabular}
    \caption{Performance on case-specific tasks for DALPHIN$_{full}$, comparing the contextual (feedback/sequential) and independent answer generation scenarios (see `VLM evaluation' in Methods). 95\% confidence intervals from bootstrapping are included in parentheses. *$P$<0.05; **$P$<0.01.}
    \label{tab:q_all_iii_contextual_independent}
\end{table}

\end{document}